\documentclass[letterpaper, 10pt, conference]{IEEEtran}
\usepackage{times}
\usepackage{xr}
\usepackage{amsmath,amssymb,mathrsfs}
\usepackage[]{algorithm2e}
\usepackage{graphicx}
\usepackage{xcolor}
\usepackage{wrapfig}
\usepackage{textcomp}
\usepackage{booktabs}
\usepackage{multirow}
\usepackage{float}

\definecolor{revision}{RGB}{203, 65, 107}
\definecolor{nicegreen}{RGB}{61, 153, 115}
\definecolor{nicepurple}{RGB}{148,0,211}

\usepackage[numbers]{natbib}
\usepackage{multicol}
\usepackage[bookmarks=true, hidelinks]{hyperref}

\begin{document}

% \includepdf[pages=-]{2024_Strgar_RSS_Rebuttal.pdf}

% paper title
% \title{Massively parallel evolution and learning in robots}
% \title{Evolving increasingly differentiable robots}
% \title{Vast evolution and learning in differentiable robots}
% \title{Massive evolution of differentiable robots}
\title{Evolution and learning in differentiable robots}

% You will get a Paper-ID when submitting a pdf file to the conference system
\author{Author Names Omitted for Anonymous Review. Paper-ID 386}

% avoiding spaces at the end of the author lines is not a problem with
% conference papers because we don't use \thanks or \IEEEmembership

\author{
\IEEEauthorblockN{Luke Strgar, David Matthews, Tyler Hummer, Sam Kriegman}
\IEEEauthorblockA{Northwestern University}
}

% \teaser{
% \centering
% % \vspace{-6pt}
% \includegraphics[trim={0 0 0 0},clip,width=\linewidth]{fig/teaser.png} \\
% \vspace{-4pt}
% \caption{teaser!
% % (\href{https://youtu.be/...}{\textcolor{blue}{\textbf{\texttt{youtu.be/...}}}}).
% } 
% \label{fig:teaser}
% \vspace{-20pt}
% }

\maketitle

\begin{abstract}
The automatic design of robots has existed for 30 years but has been constricted by serial non-differentiable design evaluations, premature convergence to simple bodies or clumsy behaviors, and a lack of sim2real transfer to physical machines.
Thus, here we employ massively-parallel differentiable simulations to rapidly and simultaneously optimize individual neural control of behavior across a large population of candidate body plans and return a fitness score for each design based on the performance of its fully optimized behavior.
Non-differentiable changes to the mechanical structure of each robot in the population---mutations that rearrange, combine, add, or remove body parts---were applied by a genetic algorithm in an outer loop of search, generating a continuous flow of novel morphologies with highly-coordinated and graceful behaviors honed by gradient descent.
This enabled the exploration of several orders-of-magnitude more designs than all previous methods, despite the fact that robots here have the potential to be much more complex, in terms of number of independent motors, than those in prior studies.
We found that evolution reliably produces ``increasingly differentiable'' robots: body plans that smooth the loss landscape in which learning operates and thereby provide better training paths toward performant behaviors.
Finally, one of the highly differentiable morphologies discovered in simulation was realized as a physical robot and shown to retain its optimized behavior.
This provides a cyberphysical platform to investigate the relationship between evolution and learning in biological systems and broadens our understanding of how a robot's physical structure can influence the ability to train policies for~it.
Videos and code at \href{https://sites.google.com/view/eldir}{\color{blue}\textbf{https://sites.google.com/view/eldir}}. 
\end{abstract}

\IEEEpeerreviewmaketitle

\begin{figure*}
    \centering
    \includegraphics[width=0.95\textwidth]{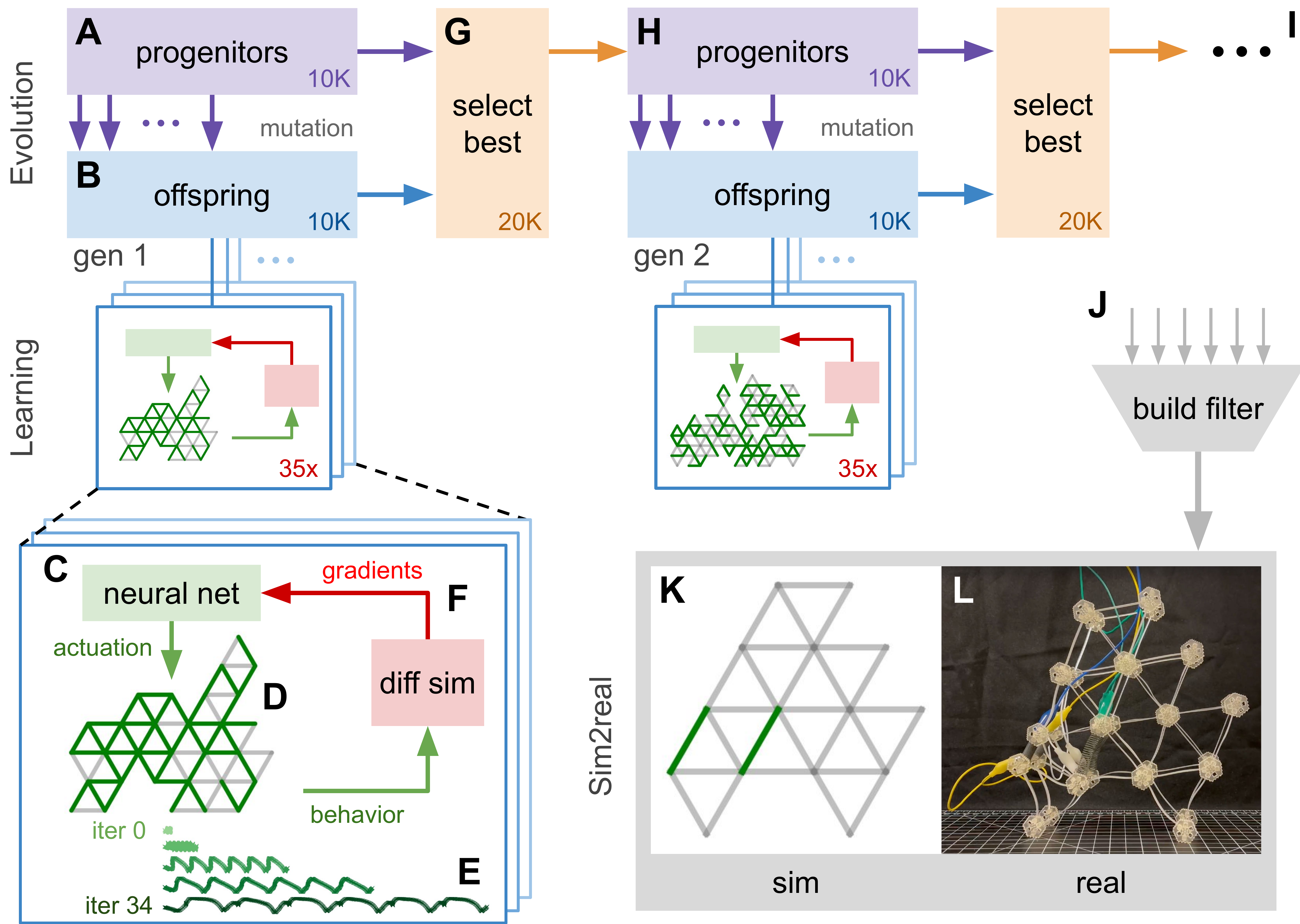}
    \caption{\textbf{Evolution and learning.}
    An initially random population of 10K progenitor robots (\textbf{A})
    produce 10K offspring (\textbf{B})
    through random morphological mutations and/or crossover,
    temporarily doubling the size of population to 20K robots.
    The 10K offspring 
    are then simultaneously trained
    in parallel differentiable simulations as follows.
    Each robot has its own
    proprioceptive neural network (\textbf{C}) 
    that coordinates the actuation of
    its motors (\textbf{D}) through differentiable simulation to produce behavior (\textbf{E}), yielding a performance score based on the net displacement of the robot in the desired direction of travel (into the right hand side of the page).
    Gradients are then propagated backward in time through the simulated behavior (\textbf{F}) and used to update the neural net's initially random synaptic weights such as to improve its performance.
    This process is repeated for 34 additional iterations of gradient descent (35 total gradient descent steps), after which
    the robot's fitness was taken to be the best performance score it achieved during training.
    Finally, selection (\textbf{G}) reduces the population back to 10K robots by deleting the worst performing robots.
    Evolution then proceeds to the next generation (\textbf{H}) and this process of design variation, parallel differentiable training, and selection is repeated 998 times for a total of 1000 generations of evolution (\textbf{I}).
    A build filter (\textbf{J}) was then used to identify the most manufacturable designs discovered in simulation; e.g.~the simulated design in \textbf{K} which was printed (\textbf{L}) and fitted with shape memory alloy springs 
    that can be repeatedly energized and cooled to generate forward locomotion.  
    }
    \label{fig:summary}
\end{figure*}

\begin{table*}[t]
\caption{\label{table:lit_review}Summary of 
% de novo 
nonparametric morphological exploration within a single run of automatic robot design.$^1$ 
% In cases in which optimization occurred on a single adaptive timescale, the number of designs explored is an upper bound.
}
\vspace{-1.25em}
\begin{center}
\begin{tabular}{lccccccc} 
    \toprule
    \textit{} 
    & \textit{}
    & \textit{}
    & \textit{Adaptive}
    & \textit{Independent} 
    & \textit{} 
    & \textit{Designs}
    & \textit{Realized} \\ 
    \textit{Author/citation} 
    & \textit{Year}
    & \textit{Material}
    & \textit{timescales}
    & \textit{motors} 
    & \textit{Pop size} 
    & \textit{explored}
    & \textit{physically} \\ 
    \midrule

    \citet{sims1994competition} & 1994 & Rigid & Single & $\sim$12  & 300 & 30K & No \\ % 100 gens

    \citet{ventrella1994explorations} & 1994 & Rigid & Single & $\sim$12 & 100 & 1K & No  \\ % 1000 crossover events between random pairs, new replaces worst.

    \citet{lipson2000automatic} & 2000 & Rigid & Single & $\sim$12 & 200 & 100K & \textbf{Yes}\\ % First generation only contains the null design; gens were 300-600; approx 10^5 evals

    \citet{komosinski2001comparison} & 2001 & Soft & Single & $\sim$5 & 200 & 60K & No \\ % 300 gens

    \citet{bongard2003evolving} & 2003 & Rigid & Single & 49 & 200 & 40K & No \\ % 200 gens; Evolution of development 50 moving parts.  2001: bongard2001repeated ; 2003:bongard2003evolving
    
    % \citet{hornby2001evolving} & 2001 & No & Rigid & 100 & 500 & \\ % L-systems ; up to 350 parts.
    % same as below

    \citet{hornby2003generative} & 2003 & Rigid & Single & \textbf{349} & 100 & 50K & \textbf{Yes} \\ % 500 gens

    \citet{miconi2006improved} & 2006 & Rigid & Single & $\sim$12 & 200 & 100K & No  \\ % 2 pops of 100 compete; 100K evaluations total 

    \citet{chaumont2007evolving} & 2007 & Rigid & Single & $\sim$3 & 300 & 30K & No  \\ % 100 gens

    % \citet{lassabe2007new} & 2007 & Rigid & Single & $\sim$12 & 100 & ? & No  \\ % gens not specified; stops when fit is good 
    % # of designs is unclear

    % \citet{miconi2008evosphere} & 2008 & Rigid & Single & $\sim$12 & 200 & ? & No \\ % continuous time; 1000 cycles of 5000 time steps. 
    % # of designs is unclear
    
    \citet{hiller2010evolving} & 2010 & Soft & Single & 2 & 20 & 25K & No \\ % 1250 gens

    \citet{krvcah2010solving} & 2010 & Rigid & Single & $\sim$12 & 300 & 45K & No  \\ % population is set to 300, with 150 generations per run.

    \citet{auerbach2011evolving} & 2011 & Rigid & Single & $\sim$20 & 150 & 75K & No  \\ % 500 gens

    \citet{lehman2011evolving} & 2011 & Rigid & Single & $\sim$12 & \textbf{1000} & 500K & No  \\ % 500 gens; Novelty search with local competition 

    \citet{hiller2012automatic} & 2012 & Soft & Single & 1 & 50 & 125K & \textbf{Yes} \\
     % 80K evaluations in Fig 6a: usually 1500-2500 gens; 75000-125000 attemps, one stopped early at 25K; we put 50K in our PNAS table...

    \citet{lessin2013open} & 2013 & Rigid & Single & $\sim$12 & 200 & 400K  & No  \\ % in thesis popsizes used are 35, 50, 100, 200; gens: 50, 60, 100, 300, 500, 1000, 2000

    \citet{cheney2013unshackling} & 2013 & Soft & Single & 2 & 30 & 30K & No  \\ % 1000 gens; No distinction btw morphology and control. \\

    \citet{cheney2014electro} & 2014 & Soft & Single & 1 & 30 & 30K & No  \\

    \citet{rieffel2014growing} & 2014 & Soft & Single & 1 & 20 & 10K & No \\ %  pop size for morphology evolution is 20; gens: 500; Either the brain or the body was held constant during evolution. Motors: "Using linear actuators with attachment vertices as in our previous two approaches would have further complicated the grammar. We chose instead to periodically vary the stiffness of the tetrahedral mesh, which results in corresponding deformations in the soft body itself."

    \citet{auerbach2014environmental} & 2014 & Rigid & Single & 2 & 150 & 75K & No \\ % Table S1. gens 500; CPPN -> marching cubes ->set of enclosed trimesh components -> the largest (most triangles) selected ->copied and reflected across the x-axis -> these two parts are connected two universal joint to a central capsule... "The two mechanical degrees of freedom of each organism are actuated by means of coupled oscillators. Each of the two oscillators is parameterized by several parameters: amplitude, period, and phase shift. These six parameters (three parameters apiece for each of the two joints) are directly encoded in the genome of the evolving organisms as floating point values so that the genome"

    \citet{brodbeck2015morphological} & 2015 & Rigid & Single & 4 & 10 & 100 & \textbf{Yes}  \\ % Table S1.3: if there are 5 blocks, mutation P(add block) = 0.

    \citet{joachimczak2016artificial} & 2016 & Soft & Single & 2 & 300 & \textbf{600K} & No \\ % population size of 300 and runs of 2000 generations

    % \citet{veenstra2017evolution} & 2017 & Rigid & Single & 19 & ? & 25K & No  \\  % asked frank to dig up pop size; not stated in paper; 20 modules max, passive seed block.

    \citet{cellucci20171d} & 2017 & Soft & Single & 2 & 100 & 30K & No  \\ % gens: 300; 1D flexible robots % PNAS table says 27000; should be 30000.

    \citet{cheney2018scalable} & 2018 & Soft & Single & 1 & 50 & 250K & No \\ % gens: 5000; Controller is evolved vector of 1000 phase offsets. motors is \textbf{1000} or 1...

    \citet{kriegman2018interoceptive} & 2018 & Soft & Single & 1 & 25 & 125K & No \\  % Controller is evolved vector of 1000 phase offsets. 
    % body, initial stifness, controller are evolved
    % stiffness develops with feedback from the environment

    \citet{corucci2018evolving} & 2018 & Soft & Single & 1 & 10 & 90K & No \\ % controller is 25 phase shifts; gens: 9000; pop size not stated. in cited code example it is 10; but that looks like a default across exps. Asked Nick, he said 10 sounds right, (they were collected quickly for an ALife extended abstract) but his default was 30 for whatever reason; Nick: "I think your guess of 10 is a good one.  I think that these results were pulled together very quickly... "

    \citet{wang2019neural} & 2019 & Rigid & \textbf{Bilevel} & $\sim$5 & 64 & 13K & No  \\ % Setc. E: 200 gens; pop count depends on number of policy training iterations and timesteps per update; Seems like min is 20 in benchmarks for which they give example of 64 species with min timesteps. Generations in table 4 also say that they can be 400 even though the results in the paper go to 200. So 64 is would seem to be a good estimate of pop size (64*200=12800). Will not change order of magnitude. Policy weight sharing between robot and its offspring. \\

    \citet{pathak2019learning} & 2019 & Rigid & Single & 5 & N/A & 1K & No   \\
    % 1200 environment steps; body is determined (self assembled) early in simulation and more or less constant during behavior
    % github repo does not contain supporting code

    \citet{kriegman2020xenobots} & 2020 & Bio & Single & 1 & 50 & 50K & \textbf{Yes} \\ 

    \citet{kriegman2020scalable} & 2020 & Soft & Single & 1 & N/A & 7K & \textbf{Yes} \\ % 6561 exhaustive search

    \citet{veenstra2020different} & 2020 & Rigid & Single & 19 & 100 & 50K & No  \\  % 500 gens; max depth of tree is 8; max modules is 20

    \citet{miras2020environmental} & 2020 & Rigid & Single & 14 & 100 & 10K & No \\ %100 gens; 15 max modules, one is head
    
    \citet{zhao2020robogrammar} & 2021 & Rigid & \textbf{Bilevel} & $\sim$20 & 1 & 2K & No  \\ % evaluate 1 design per iteration. Graph heuristic search; RL-adjacent; recursion limit is 40 which could mean many more joints but most creatures are small, none shown with more than 20
    
    \citet{kriegman2021kinematic} & 2021 & Bio & Single & 1 & 16 & 8K & \textbf{Yes} \\ % 500 gens; Self replicating machines so there are more nested designs but only count progenitors. \\
    
    \citet{kriegman2021fractals} & 2021 & Soft & Single & 2 & 16 & 5K & \textbf{Yes} \\ %325 gen; Fractal robots evaluated at three size scales, only counting basal. \\

    \citet{gupta2021embodied}  & 2021 & Rigid & \textbf{Bilevel} & 10 & 576 & 4K & No \\ % stated "less than 10 limbs" and later "at most 10 limbs". There is also a torso, so 10 motors. Fig S1: 576 initial agents at generation 0. asynchronous; \\ % 288 parallel random selections of 4 parents, best has child, then RL (512 batch; 4 policy epochs; 32 envs); 3 runs. 4000 designs total searched.

    \citet{moreno2021emerge} & 2021 & Rigid & Single & $\sim$10 & 40 & 25K & \textbf{Yes} \\ % 625 gens 

    \citet{iii2021taskagnostic} & 2021 & Rigid & Single & 9 & 24 & 1K & No  \\ % 24*60=1440;  60 gens, bodies evolved with random motor commands, then RL. They say "...under the maximum allowed number of limbs" but max limbs not stated in the paper. No head node like Gupta et al. so joints = limbs - 1. Looks like 10 limbs in the pictures. Stated in codebase max_nodes=10 which is used to set the number of limbs in 3D locomotion experiment: https://github.com/jhejna/morphology-opt/blob/7d836bf4b2f175afb2714a4e2fc791923740ad32/configs/locomotion3d/3d_tame.yaml#L14

    \citet{bhatia2021evolution} & 2021 & Soft & \textbf{Bilevel} & 25 & 25 & 1K & No  \\
    % in supplemental. pop size: 25-50; Figs 41: gen 30/40 is end of evo; 750 evals marked on xaxis of figs 3 & 40; 750/30=25

    \citet{medvet2021biodiversity} & 2021 & Soft & Single & 25 & 100 & 30K & No  \\ % distributed neural net; total evals: 30000 (so gens must be 300)

    \citet{ma2021diffaqua} & 2021 & Soft & Single & 2 & 1 & 25 & No  \\ % interpolation not de novo. \\

    \citet{yuan2022transformact} & 2021 & Rigid & Single & $\sim$14 & N/A & 1K & No \\ 
    % body designed before behavior using first few actions of policy; 
    % 5 skeleton transformations; more branchings, but 14  motors max in practice; 
    % 1K epochs; 1K unique genotypes
    % baselines compare against evolution with gens=125, popsize=20 = 2.5K

    \citet{schaff2022soft} & 2022 & Soft & Single & 8 & N/A & 50K & \textbf{Yes} \\
    % batch: 512; 1M timesteps; 20 steps per design

    \citet{van2022co} & 2022 & Soft & Single & 8 & 1 & 100 & No \\

    \citet{norstein2023effects} & 2023 & Rigid & Single & 20 & 64 & 8K & No \\ % 128 gens * 64 pop; modular robots with motors phaseshifted from global signal. They could be considered dependent.

    \citet{pigozzi2023factors} & 2023 & Soft & Single & 100 & 100 & 30K & No \\ % local neural control; https://dl.acm.org/doi/pdf/10.1145/3587101

    \citet{matthews2023efficient} & 2023 & Soft & Single & 1 & 1 & 10 & \textbf{Yes} \\ % 2816 particles; 64 muscle patches \\

    \citet{yuhn20234d} & 2023 & Soft & Single & 4 & 1 & 1K & No \\ % ITER - 2d Walker: 1.1K; 3d walker: 1.2K ; 3D rotator: 1.1K; 2d climber: 3.6K; 4 actuators, but at end they show a pic of design + loss for 2d walker with 2 and 8 actuators but no details. 

    \citet{cochevelou2023differentiable} & 2023 & Soft & Single & 2 & 1 & 200 & No \\

    \citet{li2024reinforcement} & 2024 & Soft & Single & 2 & N/A & 320K & No \\
    % 2.5K epochs
    % batch size 128

    % {\color{red}\citet{wang2023diffusebot}} &  {\color{red}2023} &  {\color{red}Soft} &  {\color{red}4} &  {\color{red}1} &  {\color{red}idk} &  {\color{red}No} \\  % this is pretrained and onyl produces blobs
    
    \textbf{This study} & {2024} & {Soft} & \textbf{Bilevel} & \textbf{453} & \textbf{10K} & \textbf{10M} & \textbf{Yes} \\
    \bottomrule
\end{tabular}
\end{center}
% \vspace{-1.75em}
\vspace{-2.5em}
\end{table*}

% \begin{figure*}
%     \centering
%     \includegraphics[width=\textwidth]{fig/zoo.pdf}
%     \caption{\textbf{A diversity of body shapes}  
%      were discovered by evolution, each with its own unique internal distribution of active (green) and passive (gray) springs. The optimized performance (displacement in the desired direction of travel) achieved by gradient based learning is denoted below each design {\color{revision} in meters (m)}. 
%     }
%     \label{fig:best-robots}
% \end{figure*}

\begin{figure*}
    \centering
    \includegraphics[width=\textwidth]{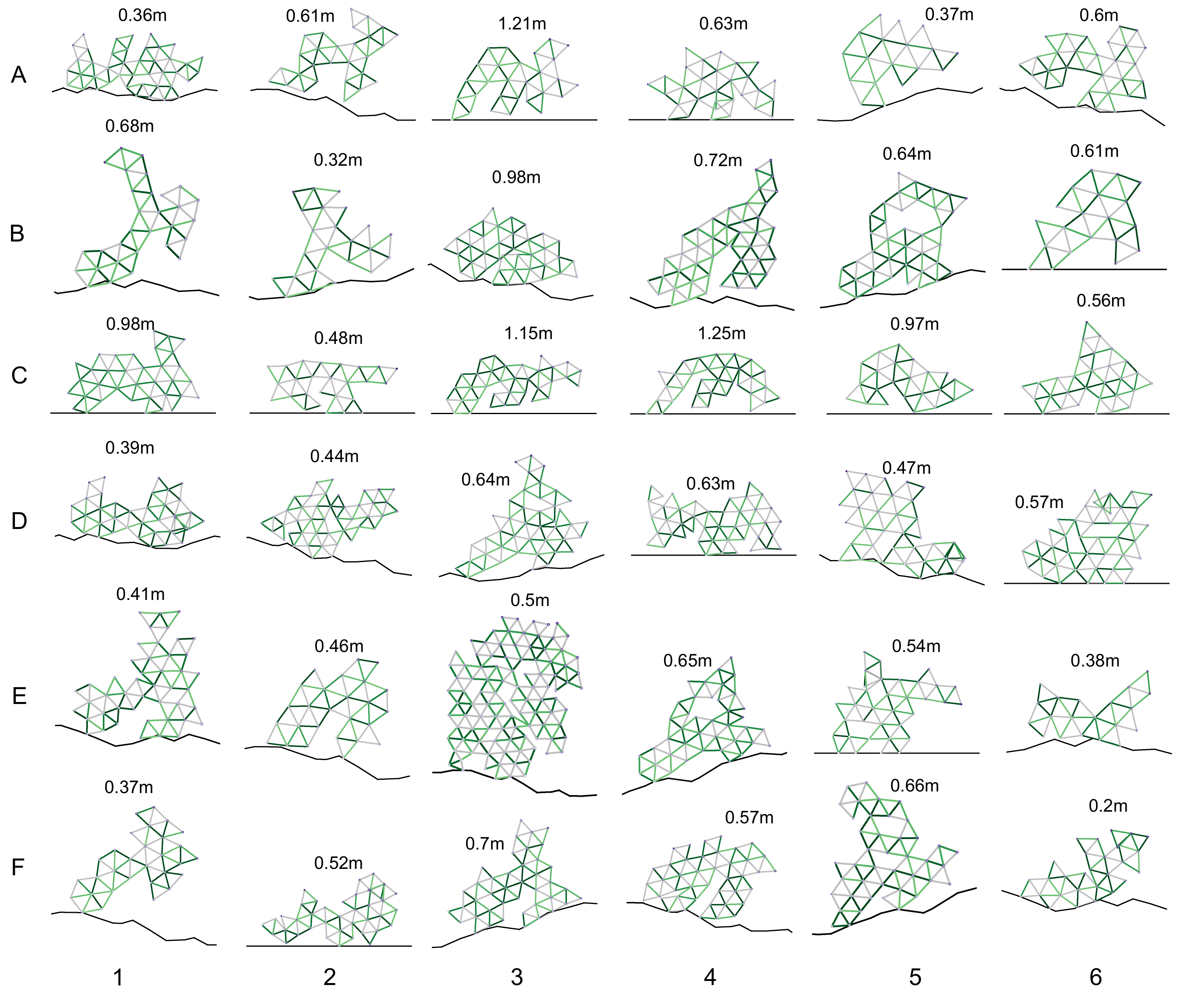}
    \caption{\textbf{A diversity of body shapes}  
     were discovered by evolution, each with its own unique internal distribution of active (green) and passive (gray) springs. 
     The optimized performance (displacement in the desired direction of travel) achieved by gradient based learning is denoted above each design in meters (m).
     % Robots 
     % % depicted here 
     % % in rows (\textbf{A-F}) and columns (\textbf{1-6})
     % evolved
     % a variety of body plans, 
     % sometimes including one, two, three, or four legs (A2, A4, C5, D6). 
     On rugged terrains 
% (Sect.~\ref{methods:terrain}) 
     bipedal body plans often emerged (A2, A6, B1, B4, B5, D3, E4, F3, F5) 
     whereas on flat terrain 
     robots often evolved three or four legs (A3, C1, C3, C4). 
     Occasionally, 
     swinging limbs that did not make contact with the ground
     were
     used to generate forward momentum
     (A4, B1, B4, E5). 
     % In some cases, what appeared to be a ``head'' was a swinging leg that did make contact with the ground during specific phases of the gait (A3, C3, C4, F6).
     Videos and code can be found at \href{https://sites.google.com/view/eldir}{\color{blue}\textbf{https://sites.google.com/view/eldir}}. 
     } 
    \label{fig:best-robots}
\end{figure*}

\section{Introduction}
\label{sec:intro}

Robots are usually trained automatically but designed manually
through 
extensive
trial and error.
The ability to automatically generate and test large numbers of 
design variants
could 
allow for new approaches to robotics
that
venture into 
previously unknown parts of design space,
revealing
forms 
and functions
that are well suited for 
their
environment
but that may not be immediately obvious to human engineers.
However, 
despite three decades of research (Table~\ref{table:lit_review}),
the
automatic design of robots
remains in its infancy,
and 
surprisingly 
few 
designs 
have been explored in simulations let alone reality.

Co-optimizing a robot's physical layout and its control policy is 
challenging 
due to the bilevel and combinatorial nature of the search problem.
Many have attempted
to collapse 
this hierarchical problem 
to a single level, 
and more recently,
to transform the discrete physical structure of a robot 
into a more continuous 
and thereby more differentiable form \cite{matthews2023efficient}. 
However, this has required imposing 
heavy constraints on either the robot's body or its brain,
and
usually results in
premature morphological convergence very early in optimization, 
effectively transforming the design problem into one of policy training for a randomly generated and morphologically simple agent~\cite{cheney2016difficulty}.

Others \cite{wang2019neural,zhao2020robogrammar,gupta2021embodied,bhatia2021evolution}
have 
respected 
the natural bilevel organization of the problem,
nesting 
the controller optimization task 
% (lower level)
inside of
the morphology optimization task 
% (upper level)
as a constraint.
Changes in robot design, like changes in animal design, occurred on a much slower (evolutionary) timescale than that of changes to neural control (learning).
However,
the
exorbitant computational cost of the inner optimization task---%
policy training using 
non-differentiable physics simulators 
to evaluate agent designs, and simulating each one forward one body part at a time---%
shackled the outer optimization task 
to a vanishingly small corner of design space:
Only a few thousand designs could be explored, 
% (5000 on avg. across these four studies) 
and those designs were relatively simple, 
possessing very few degrees of freedom, and even fewer motors.
% (15, on average).
Despite these simplifying assumptions,
no such robots have been constructed in the material world.

Differentiable simulation 
\cite{hermans2014automated,degrave2019differentiable,hu2019chainqueen,hu2019difftaichi}
has recently made possible the efficient gradient-based optimization of robots,
opening new paths 
to body-brain co-optimization.
However, 
there are only four cases reported in the literature \cite{ma2021diffaqua,matthews2023efficient,yuhn20234d,cochevelou2023differentiable}
in which 
the physical aspects of a robot's design
were allowed to change during gradient-based optimization, 
and none tackled the brain-body problem head on.
Either the
body 
was heavily constrained
(i.e.~it was interpolated between predefined basis shapes \cite{ma2021diffaqua}) 
or
there was no brain
(i.e.~actuation patterns were predefined 
and did not incorporate sensory feedback
\cite{matthews2023efficient,cochevelou2023differentiable,yuhn20234d}).
% or those patterns were directly encoded as Gaussian pulses over time without sensory feedback
All four methods
collapsed the problem to a single level
and
greedily optimized the robot's body plan
by locally descending whatever design gradient happened to be nearby its random initial conditions,
precluding exploration of the wider design space.
Moreover, they all assumed simple task environments with perfectly flat terrain. 
And though the optimized design from \citet{matthews2023efficient} was realized as a physical robot, 
it is non-trivial to generalize
gradient-based morphological search to
arbitrary hardware constraints, 
such as those imposed by 
the
relatively large, discrete and functional building blocks
% components
(motors, sensors, battery packs, wires, gears, tendons, wheels, etc.) 
that are common in embodied machines because they simplify
design, manufacture, and repair, 
but are not naturally amenable to continuous modeling.

The automatic design of robots began with a simple genetic algorithm,
and, until recently,
stochastic
gradient-free
population-based methods
were the only viable option.
Though 
this is no longer the case,
evolution
remains an attractive complement 
to gradient descent for robot design, particularly 
in the outer morphological level of search.
Indeed, 
evolutionary algorithms
are
embarrassingly parallel,
simple to implement with arbitrary constraints and multiple objectives,
and can explore 
% non-convex,
non-differentiable, rugged, sparse, and deceptive loss landscapes without becoming trapped in local optima.
They can also shine unique light on biological evolution by generating new hypotheses and 
testing those that would be otherwise difficult 
or impossible 
to test in vitro or vivo~\cite{lenski2003evolutionary,kriegman2018morphological}.
However,
the full benefits of evolutionary approaches to robot design, whatever they may be, have not been realized due to lack of scale.

One such potential benefit could be the ability to 
simultaneously optimize many robots in the same population with differing physical forms and control policies, thus realizing the desired behavior in unique ways 
(e.g.) along a Pareto front of competing objectives,
or
evenly distributed within a projection of
design space.
Indeed,
maintaining a
diverse population of suboptimal designs can in some cases be more 
useful
than 
the single 
optimal design 
for a set of explicit objectives
because 
no simulation is perfect
and some desired behaviors and 
structures are less realizable than others
(s.v.~the simulation-reality gap \cite{jakobi1995noise})
and because
it is often challenging to specify a full range of desired and undesired behaviors a priori (s.v.~the alignment problem \cite{casper2023open}).
However, 
the largest population 
used to evolve robots 
has been limited
to 
10$^3$
individuals~\cite{lehman2011evolving},
and usually contains many fewer (Table~\ref{table:lit_review}).

For several decades,
the maximum number of designs explored across an entire run of evolution 
has been on the order of 10$^5$.
However,
it is unclear 
how many unique body plans were actually visited by
brain-body co-optimization
as search was typically collapsed to a single level---a single  timescale of adaption---in which any design revision 
could have been applied 
only to the controller and left the morphology unchanged.
In such cases, 
the total number of designs explored 
was computed 
in Table~\ref{table:lit_review}
as the population size multiplied by the number of generations of evolution.
This is a rather generous upper bound. 
By this count
it was
\citet{joachimczak2016artificial}
that reported the largest prior exploration, with 600K design revisions
in a single run of optimization,
but the winning body plan ``emerge[d] in the first few hundreds of generations'' and the remainder of the run was spent training policies for it. 

%%%%%%%%%%%%%%%%%%%%%%%%%%%%%%%%%%%%%%%%%%%%%%%%
\stepcounter{footnote}\footnotetext{
It is important to note that
exploring a large subset of design space was not always the
goal of prior work in Table~\ref{table:lit_review}.
Indeed, sometimes the goal was to minimize exploration in rapid pursuit of a locally optimal design~\cite{matthews2023efficient}.
% Other times the goal was neither exploration nor exploitation but instead to understand evolution 
% robots were thus set free to compete for resources and produce offspring asynchronously in an open world
% \cite{miconi2008evosphere}.
}
% \stepcounter{footnote}\footnotetext{
% TEST
% }
%%%%%%%%%%%%%%%%%%%%%%%%%%%%%%%%%%%%%%%%%%%%%%%%

Design exploration 
has also been limited by an overall lack of morphological complexity.
Two decades ago, \citet{hornby2003generative} used a grammar-based encoding to recursively generate kinematic chains with up to 349 independently oscillating joints. 
These joints, however, were blindly actuated in an open loop without incorporating sensory information.
Since then,
no
automatically designed
robot has possessed over a hundred independent motors.
When allowed to incorporate larger numbers of actuators \cite{cheney2014electro,cheney2018scalable},
% behavior 
locomotion
was
coordinated by 
a global pacemaker
which
propagated 
a sinusoidal wave of
excitation 
% (isotropic contraction/expansion)
down the robot's entire body,
resulting in simple periodic gaits.

Here we evolve 
populations of tens of thousands of morphologically complex robots 
in parallel differentiable simulations, 
rapidly acquiring a suitable closed-loop control policy 
for each design 
using gradient descent (Fig.~\ref{fig:summary}).
Combining global (evolutionary) morphological search 
with local (gradient based) policy training 
obviates the drastic premature design convergence suffered by previous work 
and generates a steady stream of complex body plans 
with finely tuned sensorimotor coordination 
and agile behaviors along complex terrains (Fig.~\ref{fig:best-robots}). 
This allowed for exploration of 
several orders-of-magnitude 
more robot designs than any other method reported to date.
Moreover, because the simulations themselves 
are parallelized across body parts, 
robots here may contain much greater motoric complexity (i.e.~many more independent motors) than any previously evolved robots.

\section{Methods}
\label{sec:methods}

This section describes
the space of possible robot bodies (Sect.~\ref{methods:morphology}),
their neural controllers (Sect.~\ref{Methods:NeuralControl}),
the differentiable simulations (Sect.~\ref{methods:simulation}) in which they were evaluated and trained (Sect.~\ref{methods:learning}),
how different environments were modeled (Sect.~\ref{methods:terrain} and \ref{methods:ground-contact-model}),
the genetic algorithm that traverses design space (Sects.~\ref{sec:methods-mutation}, \ref{sec:methods-crossover} and \ref{sec:methods-genetic-algo}),
and the sim2real transfer of evolved robots (Sect.~\ref{sec:methods-construction}).

\subsection{Morphology.}
\label{methods:morphology}

Each robot was composed of $S$ springs connecting $M$ masses on a triangular lattice. 
The lattice was of size $A$ triangles wide and $B$ triangles tall such that the height of the lattice was approximately equal to its width. 
Specifically, triangles are assumed to be equilateral with side length of 2, thereby relating the height and width of a triangle by a factor of $\sqrt{3}$. In order to achieve a roughly square aspect ratio in the design space (lattice), A and B were set such $A \approx B\sqrt{3}$. 

\textbf{Robot geometry ($R_g$).} 
A rectangular binary mask of shape $(B,A)$ determines the presence or absence of each possible triangle, and the largest connected component was taken to be the expressed morphology. 
Triangles sharing an edge were considered connected, and triangles sharing a vertex along the vertical axis are also constituted valid connections. 
In translating from the mask to a triangular lattice, we observed it was possible for all the edges of a cell to be present while its corresponding entry in the mask was zero. 
In such cases the mask was filled appropriately. 
Henceforth we denote this mask as $R_g$ as it represents the robot's geometry.

\textbf{Internal distribution of active and passive springs ($R_s$).} 
A binary vector of length $S$, denoted $R_s$, determines which springs are capable of actuation (active) and which are not (passive) at every possible spring location on the lattice. 
The length of the vector is a result of the dimensions the grid, $A$ and $B$, and was computed automatically by enumerating all masses on the grid and counting all possible springs connecting them. 
The rest lengths of active springs may vary $\pm 10\%$ and are controlled through time using a neural-network controller as described below in the following section.

\subsection{Neural control.}
\label{Methods:NeuralControl}

\begin{figure}[]
    \centering
    \includegraphics[width=\columnwidth]{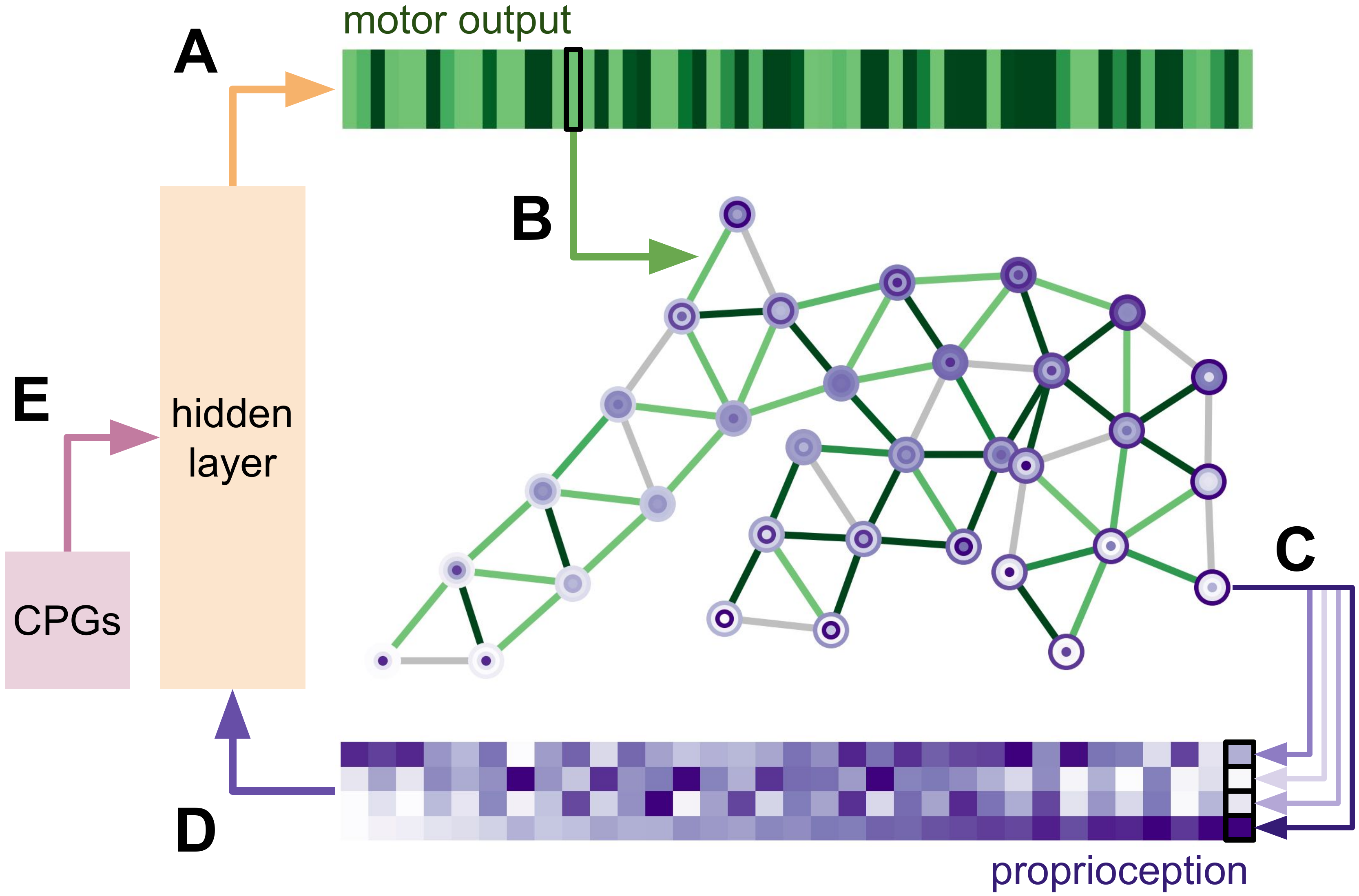}
    \caption{\textbf{Neural control of behavior.}
    Each robot is controlled by a three-layer fully-connected neural network (\textbf{A}).
    At every time step,
    motor neurons output the rest lengths of each active spring in the robot's body (\textbf{B}).
    The sensory repercussions of these actions are captured by four proprioceptors at each of the robot's masses (concentric circles in \textbf{C}),
    which feed back into the nervous system (\textbf{D}) 
    alongside central pattern generators (CPGs; \textbf{E})
    closing the control loop.
    More precisely,  
    there are 10 CPGs corresponding to 10 phase shifted sinusoidal waves;
    four proprioceptive channels track the vertical and horizontal velocity of each mass, 
    as well as the vertical and horizontal displacement of each mass relative to the robot's center of mass, during behavior.
    }
    \label{fig:neural-net}
\end{figure}

A multi-layer perceptron (MLP; Fig.~\ref{fig:neural-net}A) 
controls the actuation of each of the robot's active springs (Fig.~\ref{fig:neural-net}B)
and takes as input proprioceptive signatures across the robot's body (Fig.~\ref{fig:neural-net}C,D) as well as a set of central pattern generators (CPGs; Fig.~\ref{fig:neural-net}E).

The set of CPGs was adopted from \citet{hu2019difftaichi} and consists of 10 sine waves with a fixed angular frequency, 
$\omega = 0.08 \pi$, and equally spaced phase offsets. 
Multiple phase-shifted CPGs provide the controller an easy way to synchronize different patterns of actuation across the robot's body
at distinct times throughout their gait cycle.

At each mass in the robot's body there are four mechanosensory neurons that measure proprioception---%
the velocity and the relative displacement  of each mass,
in the horizontal and vertical dimensions. 
Displacement and velocity are measured with respect to the robot's center of mass and the initial (motionless; zero velocity) state of the robot.
Thus, the neural net of a robot with $M_r$ masses has $10 + 4*M_r$ inputs. 

The MLP contains a single hidden layer with 32 units. 
The choice of using 32 hidden units was
also adopted from \citet{hu2019difftaichi}.
Each hidden unit computes a linear combination of weights and input features plus a learnable bias before applying the $\tanh$ function to produce hidden layer activations.

Controller outputs are computed in a similar fashion with final values squashed on the interval [-1, 1]. 
Output activations are mapped one-to-one to active springs in the robot's body and used control spring actuation. 
Here, a value of -1 and 1 correspond to maximum contraction and expansion of a spring, respectively. 
Additional details of the mapping between neural controller outputs and physical simulation are provided below in Sect.~\ref{methods:simulation}.

%%%%%%%%%%%%%%%%%%%%%%%%%%%%%%%%%%%%%%%%%%%%%%%%%%%%%%%%%%%%%%
\begin{figure*}[]
    \centering
    \includegraphics[width=0.8\textwidth]{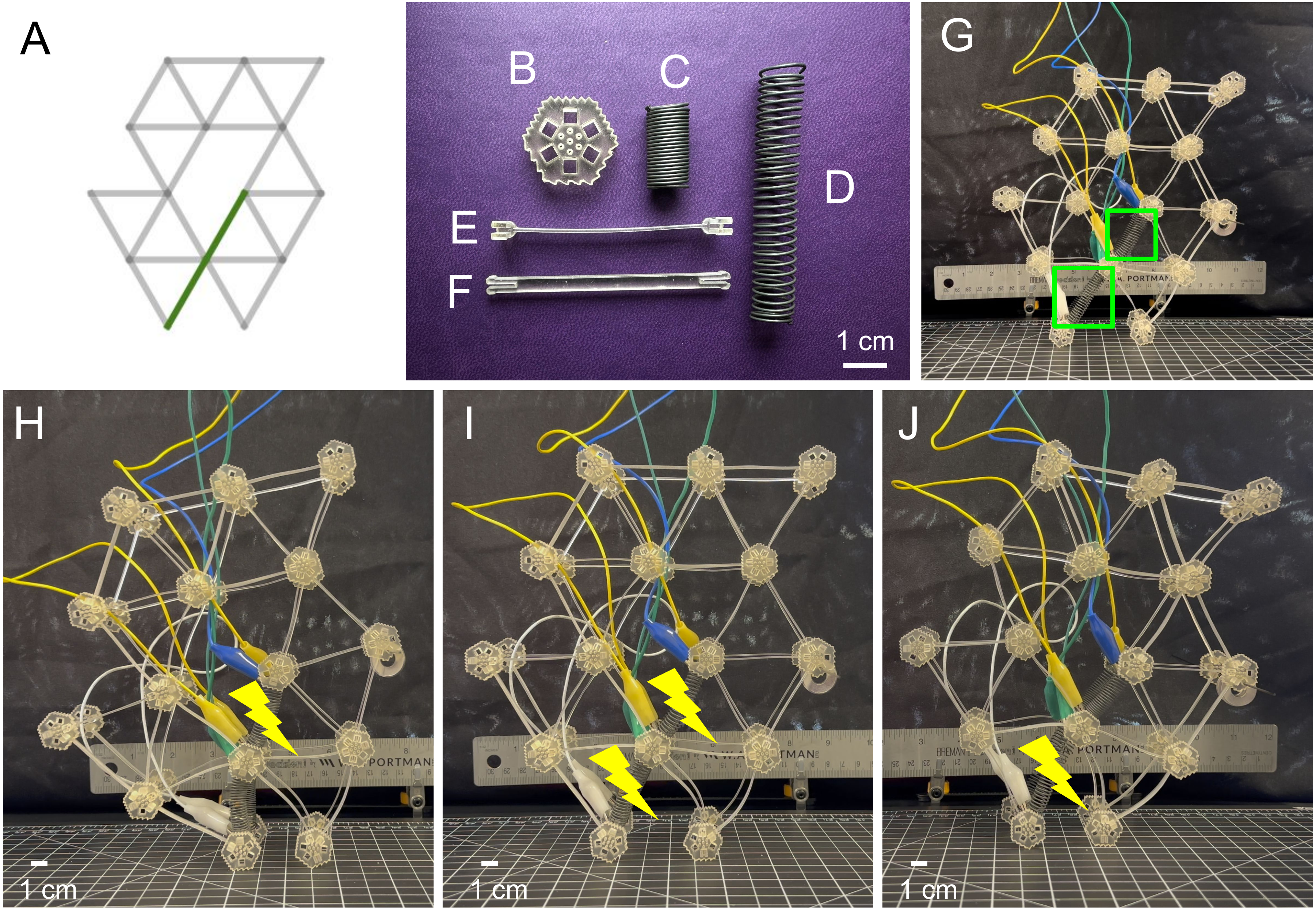}
    \caption{\textbf{Manufacture.} 
    Designs discovered in simulation (\textbf{A}) 
    were assembled from 
    3D printed hexagonal masses (\textbf{B}),
    active two-way shape memory alloy springs (\textbf{C:} contracted; \textbf{D:} expanded), 
    and passive springs (\textbf{E}). 
    Two dimensional simulated designs were transferred to 3D physical robots 
    by connecting two parallel planes of masses and springs with biplane connectors (\textbf{F}). 
    (\textbf{G-J:}) 
    The resultant robot has two active springs per planar face (four total; green boxes in G)
    that can be independently energized 
    (lightning bolts in H-J)
    to deform the robot's body.
}
    \label{fig:manufacture}
\end{figure*}

% https://docs.google.com/drawings/d/1BIbB0VvIpadmbLbQWvwgUshv396zIEMrOSifEwkcyCY/edit?usp=sharing
%%%%%%%%%%%%%%%%%%%%%%%%%%%%%%%%%%%%%%%%%%%%%%%%%%%%%%%%%%%%%%

\subsection{Simulation.}
\label{methods:simulation}
A Hookean spring-mass model was used to simulate robots. 
Masses are simulated using properties of mass and velocity as well as impulses resulting from spring actuation. 
Springs are simulated with properties including stiffness and length as well as constraints placed on maximum and minimum actuation strength. 
Actuated springs exert force on their end point masses by modulating their rest length to affect their strain and output force ($F = \Delta X * k$) for a given fixed spring stiffness $k$ and variable displacement $\Delta X$. 
In detail, each continuous neural controller (Sect.~\ref{Methods:NeuralControl}) output value is used to compute a target rest length for the corresponding active spring, and Hooke's law is used to compute the force required to achieve the aforementioned target lengths. Subsequently, continuous valued impulses acting on endpoint masses of each spring are computed and used to simulate forward dynamics (Sect.~\ref{methods:simulation}). 
Passive springs flexibly expand and contract under external forces (active springs, gravity, system dynamics) and exert impulses on their endpoint masses through tendency to preserve their fixed rest length.

We employed a GPU accelerated physics simulator (written in \textit{Taichi} \cite{hu2019difftaichi}) to simultaneously optimize neural controllers for all 10$^4$ robots in the population, evenly dividing them across 4 NVIDIA H100 SXM GPUs. 
At each time step, the actuation states and spring forces for each of the springs in each of the 10$^4$ robots was computed in parallel.

\subsection{Learning.}
\label{methods:learning}

The initially random synaptic weights of each robot's neural controller were
drawn from a scaled Xavier Normal distribution \cite{glorot2010understanding} and then 
optimized for 35 iterations of gradient descent (Fig.~\ref{fig:summary}C-F). 
This number of iterations was chosen 
because it proved sufficient in generating complex behaviors in our initial exploration of randomly generated designs.
In each iteration, the robot was simulated forward for 1000 time steps with step size $dt=0.004$, corresponding to an evaluation period of 4 seconds.
Loss was computed as the net horizontal center-of-mass displacement of the robot in the ``wrong'' direction.
That is, the negative of the net displacement of the robot's body in the desired direction of travel (i.e.~into the right hand side of the page).

Gradients were computed using the automatic differentiation features provided by Taichi~\cite{hu2019difftaichi}. 
A distinct learning rate was computed for each neural controller as a constant multiple of the inverse root sum of squares of learnable parameter gradients \cite{hu2019difftaichi}. Said parameters were then updated by applying a naive gradient descent step.

\subsection{Terrain.}
\label{methods:terrain}
% During learning (Sect.~\ref{methods:learning}) 
Robots were optimized for terrestrial locomotion across one of four distinct terrains.
The first was a perfectly flat surface line and the other three were non-uniform, ``rugged'' terrains, which were modeled by a linear piece-wise function of angled ground surfaces. 
One of these four terrains was selected \textit{a priori} and held fixed for the duration of an evolutionary run. 

Rugged terrains were generated by a sequence of surface lines that were stitched together at their endpoints. 
Each line was defined by a slope and horizontal length, both drawn from uniform distributions on a manually fixed range of values. 
Component lines were sampled and appended to the ground until reaching a maximum horizontal length of 1.25 meters.

\subsection{Ground contact model.}
\label{methods:ground-contact-model}

For flat ground
(Sect.~\ref{methods:terrain}) 
a time of impact (ToI) computation was used for mass-ground contacts. 
After advancing to the ToI, the horizontal and vertical component velocity of the mass were set to zero \cite{hu2019difftaichi}. 
These choices result in an effective ``no-slip'' friction condition for robotic locomotion. 
Such a simple model may be justified for flat terrain; however, the no-slip condition is wildly inaccurate in other conditions, endowing simulated robots with gecko-like abilities to move across simulated vertical and inverted surfaces. 

For rugged terrains 
% (Sect.~\ref{methods:terrain}) 
the ground contact and friction model was therefore modified as follows. 
As in the no-slip condition, ToI was computed and the mass was advanced to the point of contact prior to applying friction.
But here
the friction force acting on a mass was set proportional to the minimum of the mass' component velocity magnitudes pointed normal and tangent to the ground.  
The mass' resulting velocity was then the original tangent component velocity less the friction force.

We found that using the minimum of the normal and tangent velocity magnitudes to estimate friction effectively captured desired contact behavior: where the normal component velocity was small, 
little to no friction force was applied, and where tangent component velocity was small (relative to the normal component) the estimated friction force was never strong enough to flip the sign of the mass' tangent component velocity.

\subsection{Mutation.}
\label{sec:methods-mutation}

Before we describe the outer loop of morphological evolution, it is important to understand the primary variation operator: random mutation.
The geometry and active spring state of a robot were randomly modified in its descendants through operations over the binary mask, $R_g$, and vector, $R_s$ (defined in~Sect.~\ref{methods:morphology}). 
The geometry mask, $R_g$ was mutated by randomly flipping bits with probability $\frac{1}{A*B}$ such that the expected number of flips is 1. 
Here, $A$ and $B$ denote the width and height of the triangular lattice (Sect.~\ref{methods:morphology}). 
The largest connected component of the resulting mask was recomputed and compared with the original (parent) design. 
If no change had occurred the bit flipping procedure is repeated, doubling the flip probability each time. 
The spring vector $R_s$ was randomly modified in a manner identical to that $R_g$; 
however, there was no requirement to observe a change 
since the geometry containing these springs was already guaranteed to be different. 
Additionally, neither $R_s$ nor $R_g$ was permitted to be empty (containing only zeros). 

%%%%%%%%%%%%%%%%%%%%%%%%%%%%%%%%%%%%%%%%%%%%%%%%%%%%%%%%%%%%%%
\begin{figure}[!t]
    \centering
        \includegraphics[width=\columnwidth]{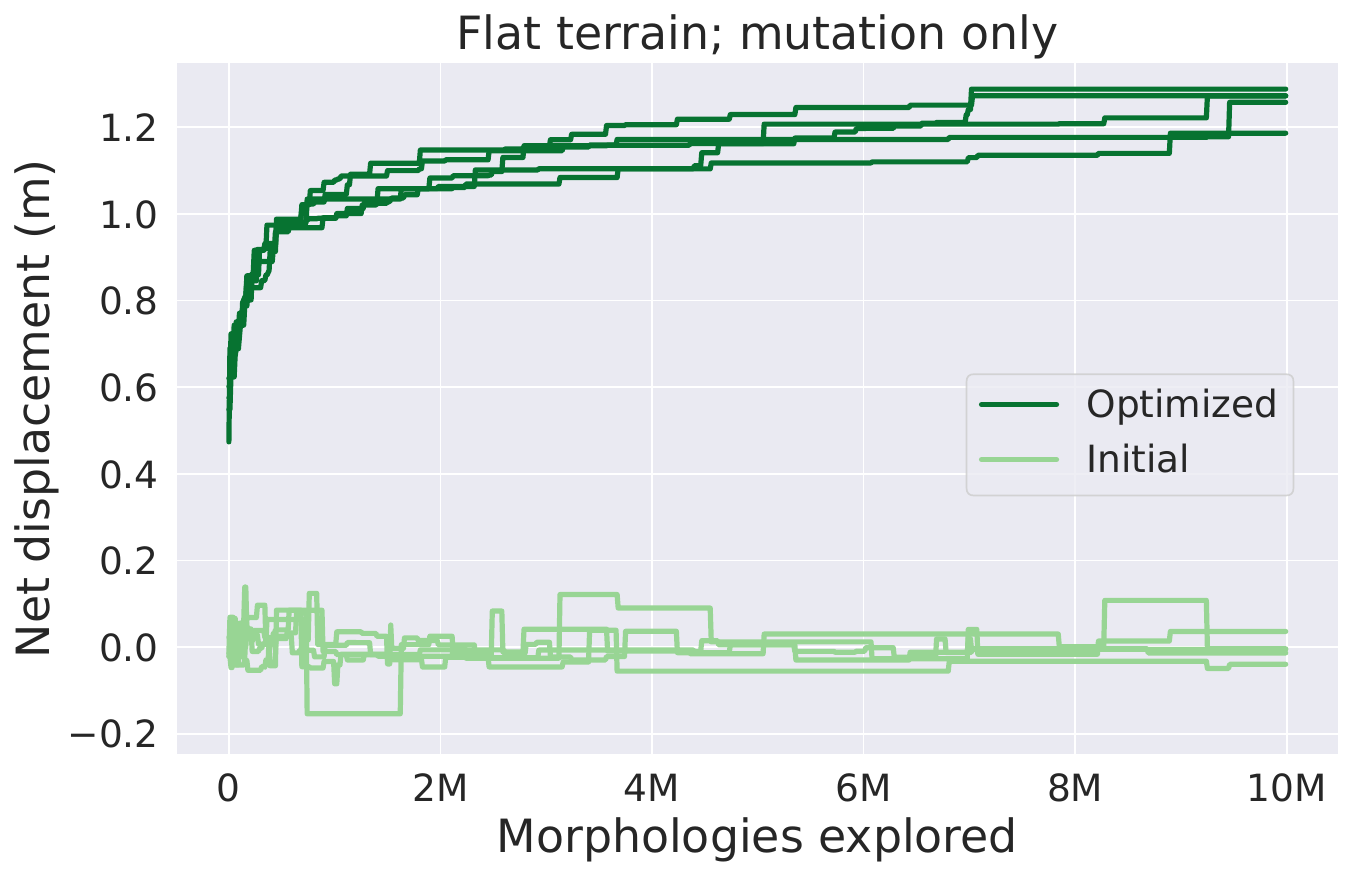}
    \caption{\textbf{Increasingly differentiable robots.}
    The initially random behavior (light green) 
    and the learned behavior (dark green)
    of the most performant design 
    are plotted 
    for five independent evolutionary trials (five pairs of light/dark lines) over flat terrain.
    Initial robot behavior 
    (at the first iteration of gradient descent) 
    produces little to no forward locomotion,
    whereas locomotive ability after learning continues to improve over evolutionary time.
    Each design was evaluated only once but may continue to survive and reproduce over many generations.
    Synaptic weights were not transmissible from parent to offspring.
    }
    \label{fig:fitness}
\end{figure}
\begin{figure}[!t]
    \centering
    \includegraphics[width=\columnwidth]{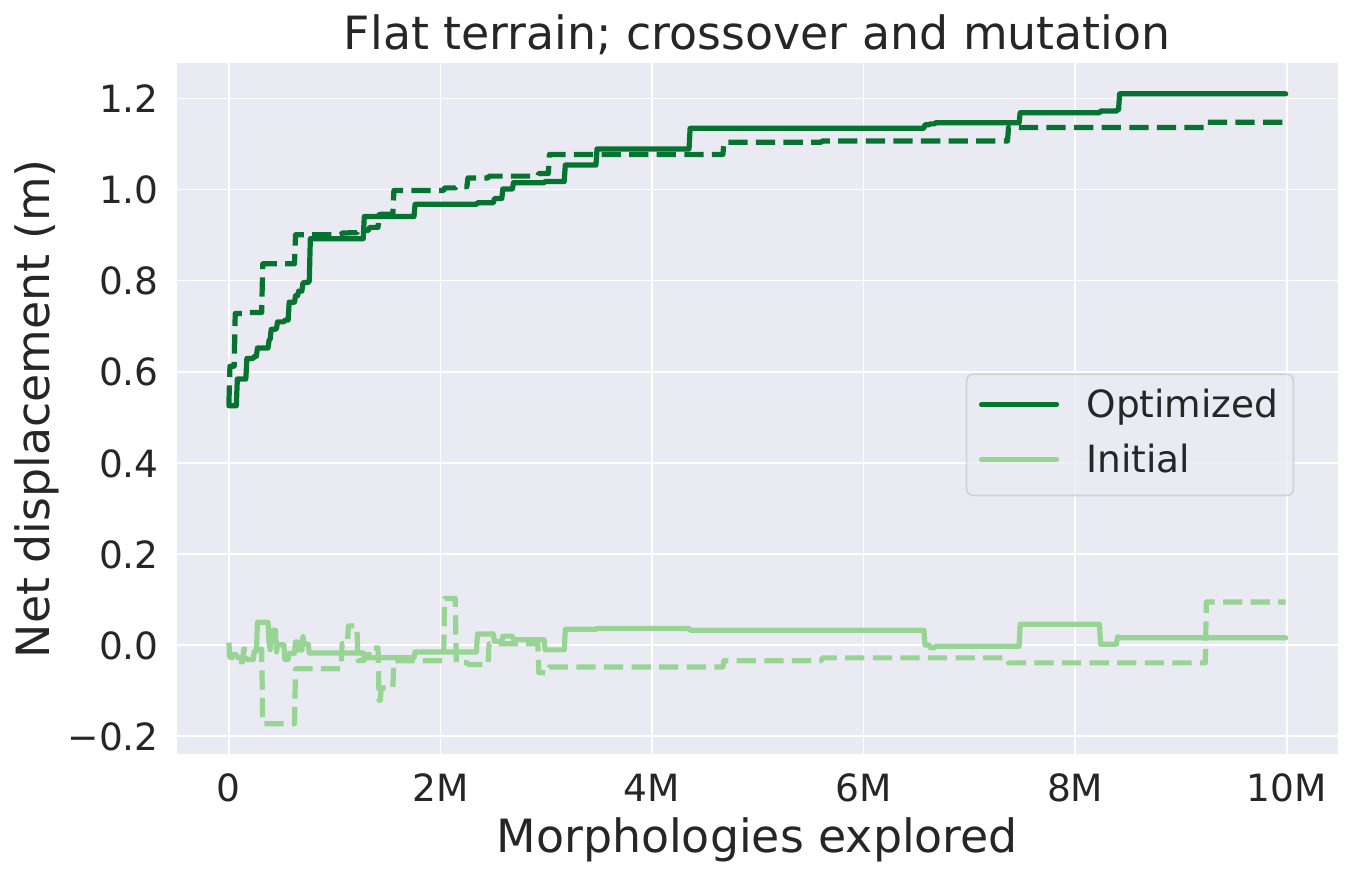}
    \caption{\textbf{With crossover.}
    The initial (light green) 
    and optimized (dark green)
    behavior of the most performant robot are plotted
    for two independent evolutionary trials in which offspring were produced through a combination of genetic recombination (crossover) and random mutation. 
    The dashed and solid pairs of lines correspond to the ``distinct'' and ``joint'' masking approaches, respectively, used in crossover (described in Section~\ref{sec:methods-crossover}).
    As in Fig.~\ref{fig:fitness}, we see an increase in the differentiability of designs as measured by their improvement during gradient based learning.
    However, these fitness curves (dark green) rise at a slightly slower rate compared to the evolutionary trials that did not employ crossover.
    }
    \label{fig:xover-fitness}
\end{figure}

\begin{figure*}[t!]
% \vspace{-1.2cm}
    \centering
    \includegraphics[width=0.96\textwidth]{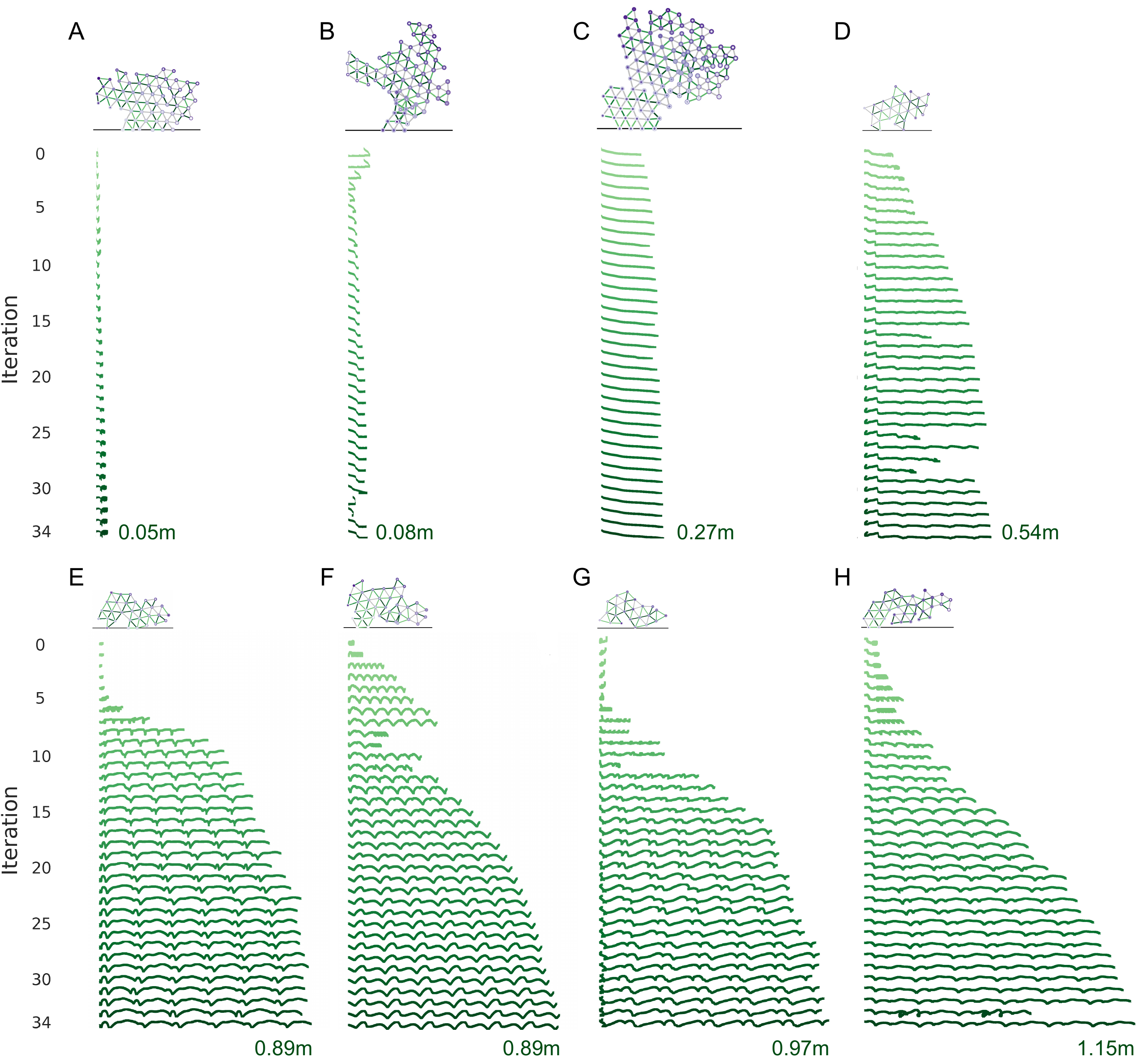}
    \caption{\textbf{Learning to walk.} 
    The behavioral trajectories of four robots that resisted learning (\textbf{A-D}) 
    and four of the highly differentiable designs that evolved (\textbf{E-H}) are shown across 35 iterations of gradient descent 
    (light to dark green CoM traces). Total horizontal displacement (fitness) is denoted to the right of (A-D) or below (E-H) the bottommost trace of each panel in meters (m). 
    With randomly initialized synaptic weights (iteration 0), all robots are more or less sessile: they have not yet learned how to effectively control their bodies.
    (A:) Early in evolution, robots often demonstrated poor differentiability: they failed to learn to walk forward (into the right hand side of the page).
    (B,C:) Perverse instantiations of the objective function---``leaning towers''---evolved to be ever taller and tip over instead of walk.
    (D:) A moderately differentiable robot that later produced more effective offspring. 
    (E-H:) Highly differentiable robots, on the other hand, show steady improvements in motility at each iteration of learning. 
    In certain cases performance can abruptly deteriorate early (F,G) or late (H) in training before recovering for a net gain. 
    The occasional deterioration in performance between iterations indicates opportunity for further development of the measure of robot differentiability.
    } 
    \label{fig:neural-trace}
\end{figure*}
\begin{figure*}[]
    \centering
    \includegraphics[width=\textwidth]{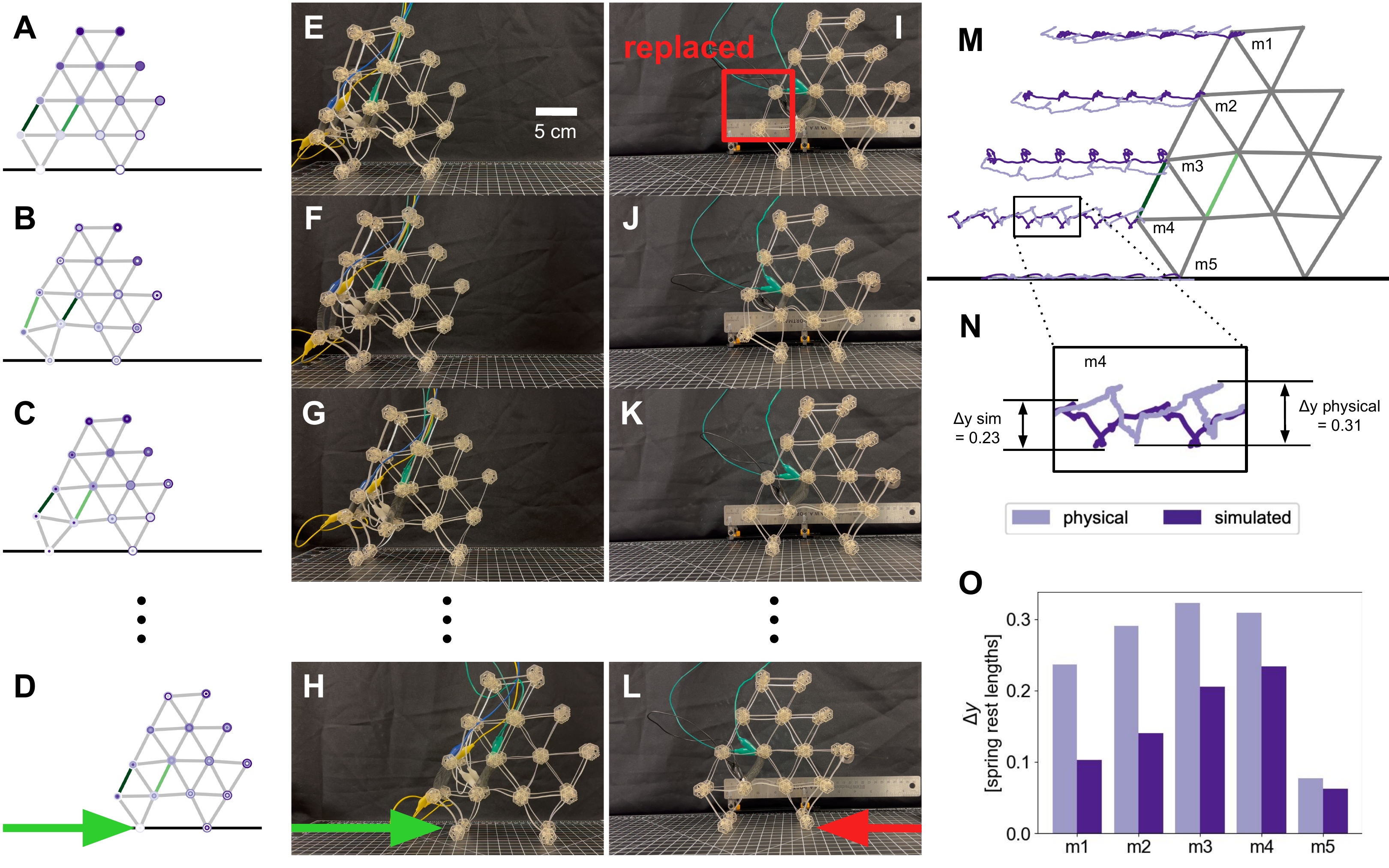}
    \caption{\textbf{Sim2real.} 
    A design discovered in simulation  
    (\textbf{A-D}) 
    locomotes over half a body length forward after 5 actuation cycles (D). 
    A robot manufactured with the same design (\textbf{E-H}) achieved a similar displacement after 5 actuation cycles (H). 
    To determine whether the robot's movement was a result of chance or due to the robot's evolved design, the design was altered by replacing one of its two motors with a passive spring (\textbf{I-L}).
    The physical instantiation of the evolved design moves in the desired direction of travel (green arrow in D and H), but, when one of its motors was replaced with a passive spring, it moved the ``wrong way''
    (i.e.~in reverse direction; red arrow in L).
    The physical trajectories (light purple traces) and the simulated trajectories (dark purple traces) of a set of masses near the actuating springs---those along the posterior edge of the design (m1-m5)---were tracked during behavior (\textbf{M}). 
    Vertical range of motion (\textbf{N}) at these parts of the physical robot's body was
    relatively consistent with the simulator's predictions: 
    the most movement occurred 
    closest to the actuating springs (m3 and m4) 
    and the least occurred at the mass in contact with the surface plane
    (\textbf{O}).
    Videos at \href{https://sites.google.com/view/eldir}{\color{blue}\textbf{https://sites.google.com/view/eldir}}.
    }
    \label{fig:sim2real}
\end{figure*}

% updated w/ motion tracking version https://docs.google.com/drawings/d/1fiDeS_QqHfKy4FrK4yq7-OFug6YTpjeYTUA-VzkU-3o/edit?usp=sharing

%  original versionhttps://docs.google.com/drawings/d/1kr6FcYhs2A7CT40JwrrxoeGSJW8c559bZjNw_Ac2hZo/edit?usp=sharing

%%%%%%%%%%%%%%%%%%%%%%%%%%%%%%%%%%%%%%%%%%%%%%%%%%%%%%%%%%%%%%

\subsection{Crossover.}
% \label{xover}
\label{sec:methods-crossover}

A second means of creating offspring robots involved combining distinct progenitor designs, typically referred to as crossover or genetic recombination. A pair of progenitor robots was randomly drawn from the population, and their geometry masks and spring vectors were fused to produce two offspring. 

Spring vectors were combined by randomly selecting, with equal probability, the value from one of the two progenitors in each position of the vector and copying that value to the offspring's spring vector. 

Two methods for combining morphology masks were explored. 
In both cases a bounding box for the largest connected component was found in each mask. 
The narrower of the two was then randomly shifted horizontally such that its width was entirely contained within the width of the wider bounding box, and an analogous operation was performed for the vertical axis. 
This ensured that a naive combination would include pieces of both progenitors.
The first method---which we term \textit{distinct masking}---randomly zeros an expected 35\% of each morphology, unions the resulting masks and takes the largest connected component as the new morphology. 
The second method---which we term \textit{joint masking}---first unions the masks then randomly zeros an expected 25\% of the result and takes the largest connected component. 
In both cases each operation is performed twice to produce a pair of offspring.
The genotype zero-masking probabilities were estimated empirically based on randomly-generated robot bodies.

\subsection{Genetic algorithm.}
\label{sec:methods-genetic-algo}
 
Evolution was seeded with a set of 10K randomly generated progenitors, 
the control policies of which were trained using gradient based learning (Sect.~\ref{methods:learning}).
Fitness was computed as the best behavior the robot achieved (absolute value of the lowest loss) across all 35 iterations of learning (Sect.~\ref{methods:learning}).
That is, the largest net horizontal center-of-mass displacement of the robot during training. 
With the population thus initialized, evolution begins (Fig.~\ref{fig:summary}A).

Every generation, each progenitor was mutated (as described in Sect.~\ref{sec:methods-mutation}) to produce an offspring (Fig.~\ref{fig:summary}B), thereby doubling the population to 20K robots.
In some experiments, crossover (detailed in Sect.~\ref{sec:methods-crossover}) was applied prior to mutation when producing offspring. 
In the experiments in which crossover occurs, it does so with an 80\% probability, 
leaving 20\% of progenitors to produce offspring through mutation only. 
The employed 80\% crossover likelihood was selected based on crossover probabilities found in the genetic programming literature, which typically range from 75\%-95\% \cite{Koza_1992_Genetic}.
In all experiments, offspring were optimized (as described in Sect.~\ref{methods:learning}) and evaluated in a manner identical to their progenitors (Fig.~\ref{fig:summary}C-F). 
Offspring and progenitors were then sorted according to fitness (Fig.~\ref{fig:summary}G) and the worst performing robots, whether progenitors or offspring, were deleted, reducing the population back to 10K robots: the next generation of progenitors (Fig.~\ref{fig:summary}H). 
This process proceeds for 10$^3$ generations, 
resulting in the exploration of a total of 10$^7$ distinct robots.

We occasionally observed numerical instability during learning 
or in the resulting mass-spring system. 
When sorting by fitness value, robots with loss values from any training iteration equaling to NaN (not a number) or infinity were considered invalid and discarded. 
To account for unstable (chaotic) mass-spring system dynamics we also filtered out robots exhibiting large absolute difference in loss value from one iteration to the next. 
Empirically, we found large deviations to be indicative of a control policy resulting in unstable body mechanics, which could generate large, artificial body displacement scores. 
Across all conducted evolutionary runs this filtering criteria eliminated less than 0.83\% of the robots evaluated and the average percentage of discarded robots in each run was 0.20\% ($\pm$ 0.29\% SD).

\subsection{Construction.}
\label{sec:methods-construction}

To ensure evolved designs could be realized as physical robots, 
evolution was constrained in a second set of experiments
to design robots within a $6\times11$ workspace and utilize at most two active springs 
(Fig. \ref{fig:manufacture}A).

Evolved robots were constructed by 3D printing three components: 
hexagonal masses (Fig. \ref{fig:manufacture}B), 
passive springs (Fig. \ref{fig:manufacture}E) 
and biplane connectors (Fig. \ref{fig:manufacture}F). 
Passive springs were fabricated with a length of 45mm and attached to the hexagonal masses to form the robot's triangular cells. 
Unlike simulated springs, passive physical springs were able to contract through bending but could not extend---relative to their fabricated length---under external forces. 
Passive physical springs exert forces on their endpoint masses through tendency toward their flattened rest length.
The 2D simulated robot was extruded into 3D, and manufactured with two identical planar faces, which were offset from each-other using biplane connectors. Connectors were designed to be rigid so as to minimize motion of component masses out of their respective planes, and every mass having fewer than six in-plane spring connections was attached to one biplane connector.

Parts were printed on a Nexa3D XiP resin printer with x45 Clear resin 
(see Table~\ref{table:resin} for resin properties).
After printing, parts were washed for two minutes while attached to the build plate in xCLEAN and subsequently one minute in IPA after being detached from the build plate.
Parts were dried and then post cured for 25 minutes.

Active springs were constructed with Nitinol two-way shape memory alloy, which had a Martensite final transition temperature of 30C and an Austenite final transition temperature of 60C.
In their high temperature state, the springs expand to be roughly 60mm long (Fig.~\ref{fig:manufacture}D), and in their cool temperature state they contract to a length of 20mm (Fig.~\ref{fig:manufacture}C).
Active springs were actuated through joule heating (11V constant voltage) and placed within a stream of room temperature air to cool.
This form of heat-based actuation 
is lightweight and provides both large deformation and high forces;
however, it is much slower than is permitted in simulation,
limiting the complexity of actuation sequences that the physical system can realize.

\section{Results}
\label{sec:results}

%%%%%%%%%%%%%%%%%%%%%%%%%%%%%%%%%%%%%%%%%%%%%%%%

\begin{figure}[b]
    \centering
    \includegraphics[width=\columnwidth]{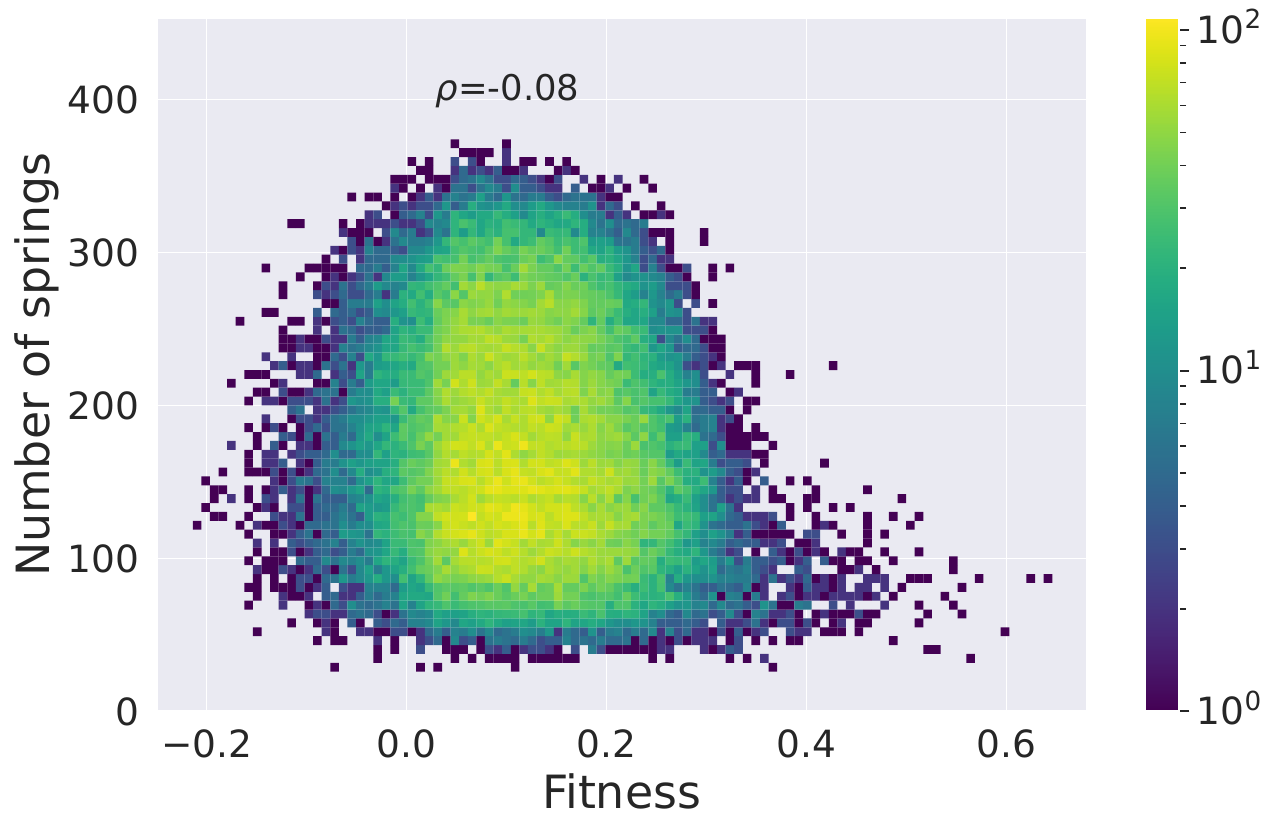}
    \caption{\textbf{The relationship between size and fitness.}
    Body size (number of springs) and fitness 
    was recorded 
    for the 70K 
    randomly-generated 
    initial progenitor robots (gen 1) 
    from the seven evolutionary conducted on flat ground terrain. 
    The Spearman rank correlation coefficient was computed as \mbox{$\rho = -0.08$}, 
    indicating little to no overall correlation between fitness and body size. 
    However,
    the lower right tail of the distribution 
    shows that the most fit randomly-generated designs posses relatively small bodies.
    }
    \label{fig:size-vs-fit}
\end{figure}
\begin{figure}[]
    \centering
    \includegraphics[width=\columnwidth]{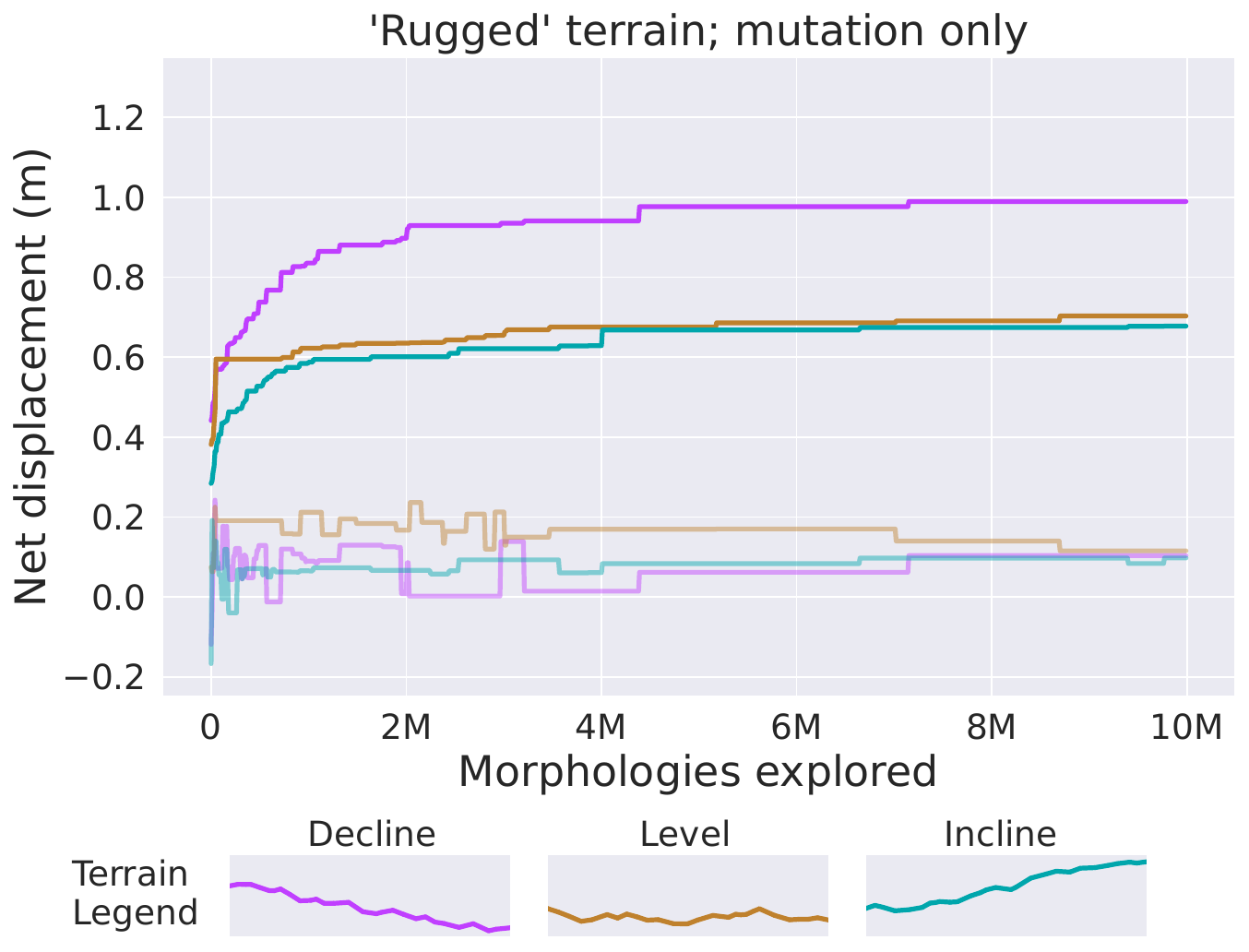}
    \caption{\textbf{Evolution in more complex environments.}
    The initial (transparent) and optimized (opaque)
    behavior of the most performant design are plotted for three independent evolutionary trials over three distinct, rugged terrains. 
    As in Figs.~\ref{fig:fitness} and \ref{fig:xover-fitness}, the increasing divergence between initial and optimized fitness demonstrate the evolution of increasingly good learners, 
    this time on more realistic (non-flat) terrains. 
    Note that fitness values on rugged terrain are not directly comparable to those achieved on flat terrain due to different ground contact and friction models (described in Sect.~\ref{methods:ground-contact-model}).
    }
    \label{fig:terrain-fitness}
\end{figure}
%%%%%%%%%%%%%%%%%%%%%%%%%%%%%%%%%%%%%%%%%%%%%%%%

\subsection{The differentiability of robots.}

\begin{figure}[b]
    \centering
    \includegraphics[width=\columnwidth]{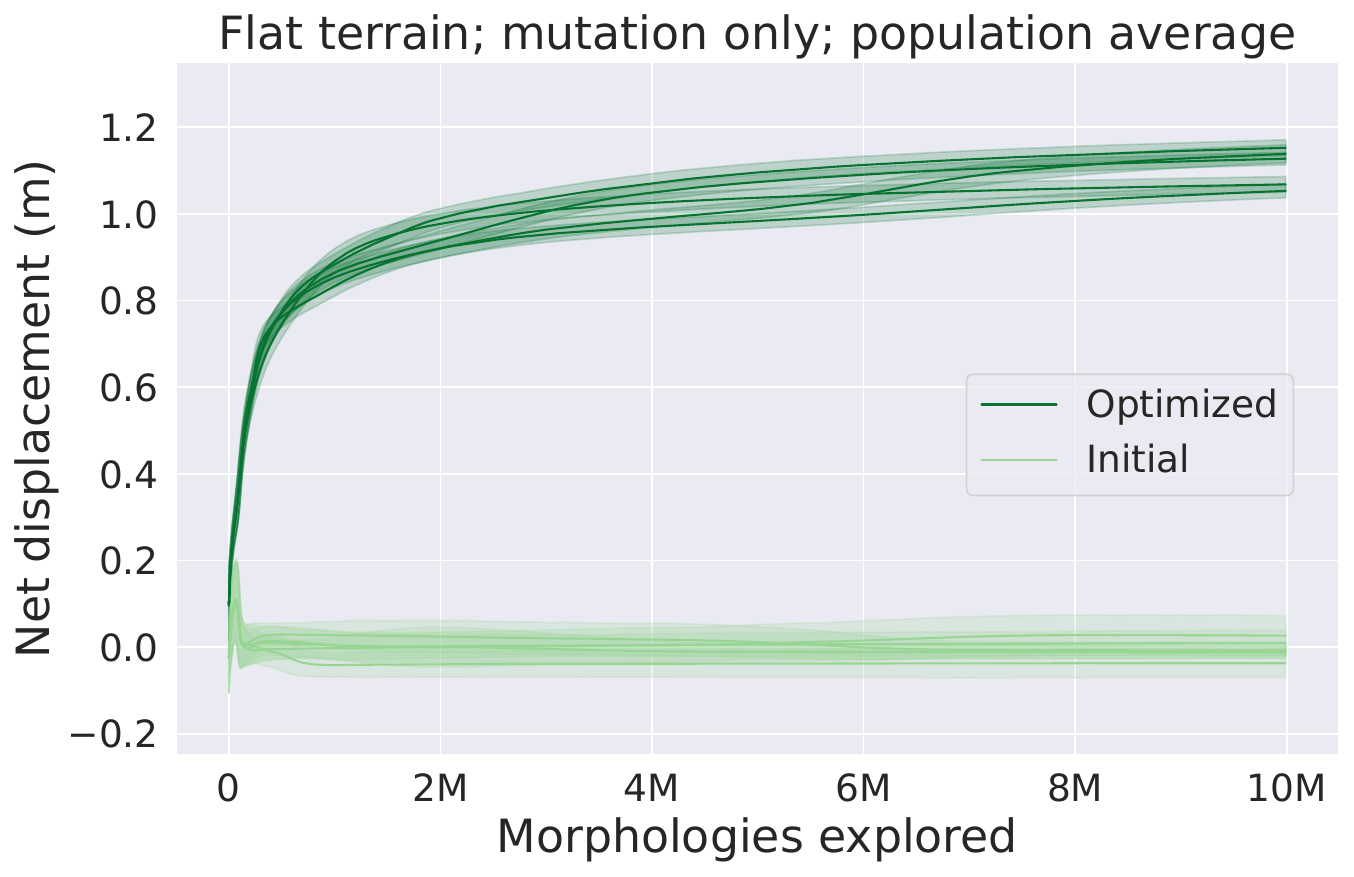}
    \caption{\textbf{Mean fitness across the population.}
    The average initial fitness (light green) 
    and 
    average optimized fitness (dark green)
    across all 10K designs in the current population
    is shown at every generation
    ($\pm$1 S.D.) 
    for the five mutation-only evolutionary trials conducted on flat terrain.
    Unlike Figs.~\ref{fig:fitness}, \ref{fig:xover-fitness}, and \ref{fig:terrain-fitness}, 
    where the fitness of the most performant design is shown at every generation, 
    here it is demonstrated 
    the population at large contains increasingly good learners and is increasingly fit overall. 
    \label{fig:population-fitness}
    }
\end{figure}

Ten independent evolutionary runs were conducted in which robots were constrained on a $13\times22$ triangular lattice (up to 453 springs and 168 masses).
For each run, the displacement of the most fit individual%
---evaluated after gradient-based learning---in each generation was plotted against the same robot's initial performance prior to learning
(Figs.~\ref{fig:fitness}, \ref{fig:xover-fitness} and \ref{fig:terrain-fitness}). 
The increasing discrepancy between fitness before and after learning indicates the discovery of morphologies
that were better suited to gradient-based control optimization in differentiable simulation.
In other words, the differentiability of the best robots in the population
increased over evolutionary time.
The same evolutionary trend was observed throughout the population as indicated by the average fitness of the fully-trained robots increasing steadily without any appreciable increases in their initial fitness prior to learning (Fig.~\ref{fig:population-fitness}).
Although synaptic 
weights were not transmitted from parents to their offspring, 
and 
there was no explicit selection pressure to increase the fitness of offspring prior to learning, 
robots could have evolved to be increasingly agnostic to synaptic weights \cite{gaier2019weight}
and thereby increasingly motile under random actuation patterns.
An example of one such design that may exist in the search space but was not found by evolution is an asymmetrical wheel that rolls forward so long as there is sufficient internal actuation~\cite{kriegman2018morphological}.

\subsection{The effect of crossover.}

To evaluate the potential for crossover to aid in morphological exploration, two of the ten independent trials included both crossover and mutation (Fig.~\ref{fig:xover-fitness}).

As in the mutation-only experiments, evolution with crossover effectively selects for robot morphologies with increasingly higher locomotive potential (Fig.~\ref{fig:xover-fitness}). 
However, in our small sample size (n=2), including crossover resulted in slightly worse performance among the most fit robots.
This may indicate that either our chosen genotypic encoding or our chosen methods of merging genotypes during crossover could be improved, as discussed in Sect.~\ref{sec:discussion}.

\subsection{Complex environments.}

To investigate the evolution of learning in more complex environments---and under a more realistic friction 
model (Sect.~\ref{methods:ground-contact-model})---we conducted three additional 
mutation-only evolutionary runs over three distinct, rugged terrains (Sect.~\ref{methods:terrain}).
The shape of these surfaces can be
seen in Figs.~\ref{fig:best-robots} and \ref{fig:terrain-fitness}.
We will refer to these environments by their net vertical tilt: 
``decline'', ``level'', and ``incline''. 
As in the case of flat terrain (Figs.~\ref{fig:fitness} and \ref{fig:xover-fitness}), 
evolution discovered body plans increasingly better at learning to move across these rugged terrains (Fig.~\ref{fig:terrain-fitness}). 
We emphasize that these robots were not augmented with any additional sensing capabilities relative to their flat terrain counterparts. 
Unsurprisingly, the fitnesses of best robots that move downhill (``decline'') superseded those of robots that moved along ``level'' terrain, which in turn superseded the fitnesses of robots that climbed up a hill (``incline'').

% \vspace{0.5em}

%%%%%%%%%%%%%%%%%%%%%%%%%%%%%%%%%%%%%%
\begin{table}[t]
\centering
\caption{\label{table:robot-size}
Size of the robots evolved on flat terrain \\ (total number of springs).}
\centering
\begin{tabular}{lcc}
\toprule
\textit{} & \textbf{Gen 1} & \textbf{Gen 1000} \\
\midrule
\textbf{All} & 176.15 ($\pm$ 67.79 SD) & 63.28 ($\pm$ 13.85 SD) \\
\textbf{Best} & 80.0 ($\pm$ 10.61 SD) & 55.0 ($\pm$ 10.65 SD) \\
\textbf{Largest} & 362.75 ($\pm$ 2.05 SD) & 84.0 ($\pm$ 16.14 SD) \\
\bottomrule
\end{tabular}
\end{table}

%%%%%%%%%%%%%%%%%%%%%%%%%%%%%%%%%%%%%%

\subsection{Morphology and fitness.}
\label{results:morphology-fitness}

When evolved for locomotion across flat terrain (Sect.~\ref{methods:terrain}) the size of the average, best, and largest robots in the population declined over the course of 1000 generations (Table~\ref{table:robot-size}).
This is due in part to a common local optimum of ``leaning towers'': 
tall, unstable bodies that oscillate until they tip over (Fig.~\ref{fig:neural-trace}B-C).
Once discovered, it is easier for evolution to simply produce ever taller offspring that fall over ever farther 
than it is for evolution to create more stable offspring with 
revised geometries and motor distributions that enable gradient descent to learn a gait yielding bona fide locomotion with net displacement comparable to that of their leaning-tower parent.

% The lack of transgenerational inheritance of neural network synaptic weights could also induce selection pressure against large robots since smaller robots with fewer motors may be easier to learn to control from scratch.
Alternatively, robots with fewer motors may be easier to learn to control, creating selection pressure for small bodies.
To test this theory, we considered the 10K randomly generated progenitors used to seed evolution in each of the seven independent trials on flat terrain. 
We ranked these 70K randomly generated robots
according to their fitness and once again according to their size (number of springs).
Spearman's rank correlation between the ranking of size and the ranking of fitness
was close to zero ($\rho=-0.08$) and statistically significant ($p<0.001$),
indicating almost no correlation between fitness and robot size (Fig.~\ref{fig:size-vs-fit}). 
Nonetheless, even if smaller robots were only marginally more fit than large robots, the population should tend towards smaller sizes, which is what we observed. 

This trend may also have been accelerated by the use of a constant mutation rate irrespective of robot size. The mutation rate could hypothetically render the discovery of adaptive mutations more difficult for larger robots where---relative to small robot---a higher number of mutations are required to introduce a new, adaptive morphological ``feature'' (e.g.~a leg) at the appropriate scale. 

While the total number of springs trended down, the fraction of active springs increased over the course of evolution (Table~\ref{table:robot-active-frac}), suggesting that greater motoric complexity (up to a point) is useful for learning effective behavior. 
Among the largest evolved robots in the population, the fraction of active springs is lower compared to the population aggregate and best, which may indicate that the employed method of neural control (Sect.~\ref{Methods:NeuralControl}) was insufficient for controlling large bodies with many independent motors.

% \vspace{0.5em}

%%%%%%%%%%%%%%%%%%%%%%%%%%%%%%%%%%%%%%
% \input{3_results_table_robot_size}
\begin{table}[t]
\centering
\caption{\label{table:robot-active-frac}
Fraction of springs that are active\\within the robots evolved on flat terrain.}
\centering
\begin{tabular}{lcc}
\toprule
\textit{} & \textbf{Gen 1} & \textbf{Gen 1000} \\
\midrule
\textbf{All} & 0.49 ($\pm$ 0.04 SD) & 0.72 ($\pm$ 0.03 SD) \\
\textbf{Best} & 0.56 ($\pm$ 0.03 SD) & 0.76 ($\pm$ 0.03 SD) \\
\textbf{Largest} & 0.50 ($\pm$ 0.03 SD) & 0.67 ($\pm$ 0.03 SD) \\
\bottomrule
\end{tabular}
\end{table}

%%%%%%%%%%%%%%%%%%%%%%%%%%%%%%%%%%%%%%

\subsection{Sim2Real.}

Unlike the simulated designs, the manufactured prototype operated with open-loop control. 
Active springs were manually actuated to imitate the actuation pattern and resulting gait of the closed-loop control policy observed in simulation following learning.

The evolved design selected for manufacture (Fig.~\ref{fig:sim2real}A)
moved forward 
0.6 body lengths over the course of five simulated actuation cycles (Fig.~\ref{fig:sim2real}A-D).
The physical instantiation of this design (Fig.~\ref{fig:sim2real}E)
moved forward at a rate of 0.8 body lengths per five actuation cycles (Fig.~\ref{fig:sim2real}E-H).

To determine if the physical robot's behavior could be attributed to the design and control discovered in simulation,
we modified the robot by removing one of its active springs.
The reduced robot completely lost the ability to locomote when actuated.
To give the reduced robot a better chance at locomoting, additional mass (9g) was added to the anterior of the robot.
When augmented in this way, the reduced robot regained the ability to locomote, albeit backwards (in the opposite direction), at rate of 0.4 body lengths over five actuation cycles (Fig.~\ref{fig:sim2real}I-L).

To gain further insight into the design's behavioral transfer, 
the trajectories of the masses along the posterior edge were tracked and plotted for both the simulated and physical bodies (Fig.~\ref{fig:sim2real}M). 
The vertical range of motion, 
$\Delta y$, 
of these mass trajectories (Fig.~\ref{fig:sim2real}N), was extracted and plotted for each of the masses (Fig.~\ref{fig:sim2real}O).
The resulting range of motion for the physical system was relatively consistent with the simulation: most motion occurred at the mass locations closest to the active springs, and the least motion occurred at the mass in contact with the surface plane.

\begin{table}[]
\centering
\caption{\label{table:resin}Material properties of the Nexa3D x45 Clear resin used to print the structure of the physical robots.}
\begin{tabular}{cll}
\toprule
\textit{ASTM code} & \textit{Property} & \textit{Value} \\
\midrule
D638 & Tensile Modulus & 1600 MPa \\
D638 & Ultimate Tensile Strength & 52 MPa \\
D638 & Tensile Elongation at Break &  12\% \\
D790 & Flex Modulus &  2100 MPa \\
D790 & Flex Strength &  95 MPa \\
D256 & Notched Izod &  19 J/m \\
D570 & Water Absorption &  6\% \\
D2240 & Hardness Shore D &  85 \\
\bottomrule
\end{tabular}
% \vspace{-1em}
\end{table}

\vspace{0.5em}

\section{Discussion}
\label{sec:discussion}

The astonishing variety and complexity of animals 
that inspires roboticists today
was
designed automatically 
% over hundreds of millions of years 
by a simple yet massively-parallel genetic algorithm. 
In this paper, we demonstrated that a similar evolution may be possible in robots
% over much shorter timescales
if the bottleneck of serial non-differentiable evaluations 
is removed and
replaced by a much wider, differentiable channel.
However, there were several limitations of our initial approach that may be overcome by future work. 
What follows is a list of ``problems'' (they may, after all, turn out to be false problems) and potential solutions to each one.
We hope that outlining problem-solution pairs in this way inspires others to engage creatively with the brain-body problem in robots
and ultimately banish the technological obstacles that stand in the way of autonomous robots with morphological complexity and behavioral competency 
rivaling that of their living counterparts.

\subsection{Limitations of differentiable simulation.}

\begin{itemize}  

\item \textbf{Problem:}
Space complexity limits the number of time steps through which gradients can be traced, since space complexity grows linearly with step count.\\

\item \textbf{Solution:}
Space complexity can be traded for computational complexity through checkpointing,~in~which~a subset of simulation time steps are saved 
and the intermediate steps 
are regenerated as needed during the backward pass.
For more details see, e.g., Appx.~D.2 of~\cite{hu2019difftaichi}.
\\

\item \textbf{Problem:} Maintaining accurate and stable gradients over long periods of simulation time
can be challenging due to the accumulation of floating point errors,
as well as 
vanishing and exploding gradients.
\\

\item \textbf{Solution:} Objective functions could be devised that depend on smaller windows of simulation time and thus
do not require backpropagating to the initial time step \cite{xu2021accelerated}.

\end{itemize}

\subsection{Limitations of the encoding.}
\label{sec:discuss-encoding}

\begin{itemize}  

\item \textbf{Problem:}
Encoding the phenotypic
details of every body part
directly
into the genotype
scales unfavorably 
with morphological size and complexity.
\\

\item \textbf{Solution:}
Indirect encodings such as 
grammars \cite{hornby2003generative}, 
directed graphs \cite{sims1994competition}, 
pattern-producing networks \cite{cheney2013unshackling},
and others \cite{veenstra2020different}
can be used to compress phenotypic complexity into a compact representation.

\end{itemize}

\subsection{Limitations of the genetic algorithm.} 

\begin{itemize}

\item
\textbf{Problem:} 
Neural structures of the robots in this paper 
were not inheritable.
Thus
any good behaviors discovered by progenitors during their lifetime
(i.e.~through gradient based learning)
had to be rediscovered in offspring ab~initio,
which precluded genetic assimilation through Baldwinian effects~\cite{kriegman2018morphological,hinton1987learning},
and
which may have limited the robots' behavioral complexity.
\\

\item
\textbf{Solution:} 
Neural architectures and synaptic weights could be transmitted transgenerationally and co-optimized during the lifetime of a robot.
Intergenerational weight sharing could also allow for ``culture'' to emerge through which ``good tricks'' are rapidly transmitted across the population.

\end{itemize}

\subsection{Limitations of the learning algorithm.} 

\begin{itemize}

\item
\textbf{Problem:} Our naive gradient-based learning procedure is not guaranteed to discover the globally optimal controller for a given morphology. 
As a result, good body designs may be discarded due to sub-optimal control optimization.
\\

\item
\textbf{Solution:} Although there is no known, provably optimal gradient based-learning scheme, many additional steps can be taken to improve traversal of the learning loss landscape and reduce the likelihood of obtaining substantially sub-optimal performance. These include, but are not limited, to adaptive learning algorithms \cite{adamkingma}.

\end{itemize}

\subsection{Limitations of a static environment.} 

\begin{itemize}

\item
\textbf{Problem:} The conducted evolutionary trials appear to be converging to highly-specialized body plans for 
the simple and static
environment and selection pressures.
\\

\item
\textbf{Solution:} 
Morphological complexity may be correlated with environmental  complexity \cite{auerbach2014environmental},
which could include other agents \cite{nolfi1998coevolving}
and competition for resources \cite{sims1994competition}.
Additional niches
may be created
in the environment 
by
introducing additional search objectives
and/or
explicitly rewarding diversity \cite{lehman2011evolving}. 

\end{itemize}

\subsection{Limitations of the variation operators.} 

\begin{itemize}

\item
\textbf{Problem:} 
Genetic recombination (crossover)
is believed to serve an adaptive function in animals \cite{otto2009evolutionary};
however,
our naive implementation of crossover 
(as described in Sect.~\ref{sec:methods-crossover})
seems to have slightly inhibited evolutionary innovation 
as indicated by 
a ``deceleration''
of the fitness curves of independent evolutions which employed crossover (Fig.~\ref{fig:xover-fitness})
compared to those of evolutions which did not (Fig.~\ref{fig:fitness}).
\\

\item
\textbf{Solution:} 
Methods for recombining neural structures \cite{stanley2002NEAT}
may be adapted for morphological structures under an appropriate genetic encoding (see Sect.~\ref{sec:discuss-encoding}).

\end{itemize}

\subsection{Limitations of the actuators.}

\begin{itemize}

\item \textbf{Problem:} Two way shape memory alloy springs are not constrained to expand along a linear path. Thus, as the active springs expand they may bend, resulting in a non-linear relationship between spring length and force exerted on its endpoints.
\\

\item \textbf{Solution:}  Alternate actuation mechanisms could be utilized which may allow for more precise control over actuation, and scale more readily to larger numbers of independent actuators in the physical robot. Some examples include tendon driven contractile actuators, pneumatic actuators, lead screw or solenoid based linear actuators.
\\

\item \textbf{Problem:} Uneven cooling of the spring actuators led to out-of-plane motion response not accounted for in simulation.
\\

\item \textbf{Solution:}  Adding uniform airflow across the entire body, or changing to alternate classes of actuators could further improve strictly planar motion.  
\end{itemize}

\subsection{Limitations of sim2real.}

\begin{itemize}
\item \textbf{Problem:} Physical robots were not able to actuate with the same strength and speed as their simulated counterparts.
\\

\item \textbf{Solution:} An iterative process of physical system characterization and simulation development could result in a tighter cyber-physical coupling. 
\\
\end{itemize}

\section{Conclusion}

Thirty years ago, the very first studies to automatically design robots 
did so by
evolving populations of a few hundred morphologically-simple robots, each composed of a dozen or so moving parts.
Since then the field has experienced a long period of stasis;
despite exponential increases in computational power 
and the embarrassingly parallel nature of evolutionary algorithms, 
contemporary work still features comparable
population sizes
and 
low sensorimotor complexity (Table~\ref{table:lit_review}). 
This lack of progress can largely be attributed to the inefficiencies of the serialized non-differentiable simulations used to evaluate design variants.

In this paper we unified evolutionary robotics with differentiable policy training and, in doing so, enhanced the scale of the former by orders of magnitude. 
Our results demonstrate a synchronicity between evolutionary morphological search and gradient based controller optimization wherein body plans with increasing fitness potential---by way of learning---are discovered. 
To demonstrate the power of joining evolution and gradient based learning at scale, we intentionally chose the most simple, ``vanilla'' algorithmic components wherever possible.
Despite these simplifications, we observed the emergence of millions of novel robots with complex anatomical forms (body shapes, musculature, and proprioceptors) that elegantly coordinated their movement across interesting terrains.
In this light, these results both highlight the power of joining large-scale evolution with differentiable training 
and suggest the potential for further advances with this framework.

\section{Code}
\label{sec:code}

% \noindent
Supporting code is available at
\href{https://github.com/lstrgar/ELDiR}{\color{blue}\textbf{github.com/lstrgar/ELDiR}}.

\section*{Acknowledgements}

This research was supported by
Schmidt Sciences AI2050 grant G-22-64506
and
Templeton World Charity Foundation award no.~20650.

\bibliographystyle{plainnat}
\bibliography{main}

\begin{thebibliography}{66}
\providecommand{\natexlab}[1]{#1}
\providecommand{\url}[1]{\texttt{#1}}
\expandafter\ifx\csname urlstyle\endcsname\relax
  \providecommand{\doi}[1]{doi: #1}\else
  \providecommand{\doi}{doi: \begingroup \urlstyle{rm}\Url}\fi

\bibitem[Auerbach and Bongard(2011)]{auerbach2011evolving}
Joshua~E Auerbach and Josh~C Bongard.
\newblock Evolving complete robots with {CPPN-NEAT}: {T}he utility of recurrent connections.
\newblock In \emph{Proceedings of the Genetic and Evolutionary Computation Conference (GECCO)}, pages 1475--1482, 2011.

\bibitem[Auerbach and Bongard(2014)]{auerbach2014environmental}
Joshua~E Auerbach and Josh~C Bongard.
\newblock Environmental influence on the evolution of morphological complexity in machines.
\newblock \emph{PLoS Computational Biology}, 10\penalty0 (1):\penalty0 e1003399, 2014.

\bibitem[Bhatia et~al.(2021)Bhatia, Jackson, Tian, Xu, and Matusik]{bhatia2021evolution}
Jagdeep Bhatia, Holly Jackson, Yunsheng Tian, Jie Xu, and Wojciech Matusik.
\newblock Evolution gym: A large-scale benchmark for evolving soft robots.
\newblock \emph{Advances in Neural Information Processing Systems (NeurIPS)}, 34:\penalty0 2201--2214, 2021.

\bibitem[Bongard and Pfeifer(2003)]{bongard2003evolving}
Josh Bongard and Rolf Pfeifer.
\newblock Evolving complete agents using artificial ontogeny.
\newblock In \emph{Morpho-functional machines: The new species: Designing embodied intelligence}, pages 237--258, 2003.

\bibitem[Brodbeck et~al.(2015)Brodbeck, Hauser, and Iida]{brodbeck2015morphological}
Luzius Brodbeck, Simon Hauser, and Fumiya Iida.
\newblock Morphological evolution of physical robots through model-free phenotype development.
\newblock \emph{PloS ONE}, 10\penalty0 (6):\penalty0 e0128444, 2015.

\bibitem[Casper et~al.(2023)Casper, Davies, Shi, Gilbert, Scheurer, Rando, Freedman, Korbak, Lindner, Freire, Wang, Marks, Segerie, Carroll, Peng, Christoffersen, Damani, Slocum, Anwar, Siththaranjan, Nadeau, Michaud, Pfau, Krasheninnikov, Chen, Langosco, Hase, Biyik, Dragan, Krueger, Sadigh, and Hadfield-Menell]{casper2023open}
Stephen Casper, Xander Davies, Claudia Shi, Thomas~Krendl Gilbert, J{\'e}r{\'e}my Scheurer, Javier Rando, Rachel Freedman, Tomasz Korbak, David Lindner, Pedro Freire, Tony~Tong Wang, Samuel Marks, Charbel-Raphael Segerie, Micah Carroll, Andi Peng, Phillip Christoffersen, Mehul Damani, Stewart Slocum, Usman Anwar, Anand Siththaranjan, Max Nadeau, Eric~J Michaud, Jacob Pfau, Dmitrii Krasheninnikov, Xin Chen, Lauro Langosco, Peter Hase, Erdem Biyik, Anca Dragan, David Krueger, Dorsa Sadigh, and Dylan Hadfield-Menell.
\newblock Open problems and fundamental limitations of reinforcement learning from human feedback.
\newblock \emph{Transactions on Machine Learning Research}, 2023.

\bibitem[Cellucci et~al.(2017)Cellucci, MacCurdy, Lipson, and Risi]{cellucci20171d}
Daniel Cellucci, Robert MacCurdy, Hod Lipson, and Sebastian Risi.
\newblock 1d printing of recyclable robots.
\newblock \emph{IEEE Robotics and Automation Letters}, 2\penalty0 (4):\penalty0 1964--1971, 2017.

\bibitem[Chaumont et~al.(2007)Chaumont, Egli, and Adami]{chaumont2007evolving}
Nicolas Chaumont, Richard Egli, and Christoph Adami.
\newblock Evolving virtual creatures and catapults.
\newblock \emph{Artificial Life}, 13\penalty0 (2):\penalty0 139--157, 2007.

\bibitem[Cheney et~al.(2013)Cheney, MacCurdy, Clune, and Lipson]{cheney2013unshackling}
Nick Cheney, Robert MacCurdy, Jeff Clune, and Hod Lipson.
\newblock Unshackling evolution: Evolving soft robots with multiple materials and a powerful generative encoding.
\newblock In \emph{Proceedings of the Conference on Genetic and Evolutionary Computation (GECCO)}, pages 167--174. ACM, 2013.

\bibitem[Cheney et~al.(2014)Cheney, Clune, and Lipson]{cheney2014electro}
Nick Cheney, Jeff Clune, and Hod Lipson.
\newblock Evolved electrophysiological soft robots.
\newblock In \emph{Proceedings of the Conference on Artificial Life (ALife)}, volume~14, pages 222--229, 2014.

\bibitem[Cheney et~al.(2016)Cheney, Bongard, SunSpiral, and Lipson]{cheney2016difficulty}
Nick Cheney, Josh Bongard, Vytas SunSpiral, and Hod Lipson.
\newblock On the difficulty of co-optimizing morphology and control in evolved virtual creatures.
\newblock In \emph{Proceedings of the Artificial life Conference (ALIFE)}, pages 226--233. MIT Press, 2016.

\bibitem[Cheney et~al.(2018)Cheney, Bongard, SunSpiral, and Lipson]{cheney2018scalable}
Nick Cheney, Josh Bongard, Vytas SunSpiral, and Hod Lipson.
\newblock Scalable co-optimization of morphology and control in embodied machines.
\newblock \emph{Journal of The Royal Society Interface}, 15\penalty0 (143):\penalty0 20170937, 2018.

\bibitem[Cochevelou et~al.(2023)Cochevelou, Bonner, and Schmidt]{cochevelou2023differentiable}
Fran\c{c}ois Cochevelou, David Bonner, and Martin-Pierre Schmidt.
\newblock Differentiable soft-robot generation.
\newblock In \emph{Proceedings of the Genetic and Evolutionary Computation Conference (GECCO)}, page 129–137. ACM, 2023.

\bibitem[Corucci et~al.(2018)Corucci, Cheney, Giorgio-Serchi, Bongard, and Laschi]{corucci2018evolving}
Francesco Corucci, Nick Cheney, Francesco Giorgio-Serchi, Josh Bongard, and Cecilia Laschi.
\newblock Evolving soft locomotion in aquatic and terrestrial environments: Effects of material properties and environmental transitions.
\newblock \emph{Soft Robotics}, 5\penalty0 (4):\penalty0 475--495, 2018.

\bibitem[Degrave et~al.(2019)Degrave, Hermans, Dambre, et~al.]{degrave2019differentiable}
Jonas Degrave, Michiel Hermans, Joni Dambre, et~al.
\newblock A differentiable physics engine for deep learning in robotics.
\newblock \emph{Frontiers in Neurorobotics}, page~6, 2019.

\bibitem[Gaier and Ha(2019)]{gaier2019weight}
Adam Gaier and David Ha.
\newblock Weight agnostic neural networks.
\newblock In \emph{Advances in Neural Information Processing Systems (NeurIPS)}, volume~32, 2019.

\bibitem[Glorot and Bengio(2010)]{glorot2010understanding}
Xavier Glorot and Yoshua Bengio.
\newblock Understanding the difficulty of training deep feedforward neural networks.
\newblock In \emph{Proceedings of the International Conference on Artificial Intelligence and Statistics}, pages 249--256. JMLR, 2010.

\bibitem[Gupta et~al.(2021)Gupta, Savarese, Ganguli, and Fei-Fei]{gupta2021embodied}
Agrim Gupta, Silvio Savarese, Surya Ganguli, and Li~Fei-Fei.
\newblock Embodied intelligence via learning and evolution.
\newblock \emph{Nature Communications}, 12\penalty0 (1):\penalty0 1--12, 2021.

\bibitem[Hejna~III et~al.(2021)Hejna~III, Abbeel, and Pinto]{iii2021taskagnostic}
Donald~J Hejna~III, Pieter Abbeel, and Lerrel Pinto.
\newblock Task-agnostic morphology evolution.
\newblock In \emph{Proceedings of the International Conference on Learning Representations (ICLR)}, 2021.

\bibitem[Hermans et~al.(2014)Hermans, Schrauwen, Bienstman, and Dambre]{hermans2014automated}
Michiel Hermans, Benjamin Schrauwen, Peter Bienstman, and Joni Dambre.
\newblock Automated design of complex dynamic systems.
\newblock \emph{PloS ONE}, 9\penalty0 (1):\penalty0 e86696, 2014.

\bibitem[Hiller and Lipson(2010)]{hiller2010evolving}
Jonathan Hiller and Hod Lipson.
\newblock Evolving amorphous robots.
\newblock In \emph{Conference on Artificial Life (ALife)}, pages 717--724, 2010.

\bibitem[Hiller and Lipson(2012)]{hiller2012automatic}
Jonathan Hiller and Hod Lipson.
\newblock Automatic design and manufacture of soft robots.
\newblock \emph{IEEE Transactions on Robotics}, 28\penalty0 (2):\penalty0 457--466, 2012.

\bibitem[Hinton and Nowlan(1987)]{hinton1987learning}
Geoffrey~E Hinton and Steven~J Nowlan.
\newblock How learning can guide evolution.
\newblock \emph{Complex Systems}, 1:\penalty0 495--502, 1987.

\bibitem[Hornby et~al.(2003)Hornby, Lipson, and Pollack]{hornby2003generative}
Gregory~S Hornby, Hod Lipson, and Jordan~B Pollack.
\newblock Generative representations for the automated design of modular physical robots.
\newblock \emph{IEEE Transactions on Robotics and Automation}, 19\penalty0 (4):\penalty0 703--719, 2003.

\bibitem[Hu et~al.(2019)Hu, Liu, Spielberg, Tenenbaum, Freeman, Wu, Rus, and Matusik]{hu2019chainqueen}
Yuanming Hu, Jiancheng Liu, Andrew Spielberg, Joshua~B Tenenbaum, William~T Freeman, Jiajun Wu, Daniela Rus, and Wojciech Matusik.
\newblock Chainqueen: A real-time differentiable physical simulator for soft robotics.
\newblock In \emph{Proceedings of the International Conference on Robotics and Automation (ICRA)}, pages 6265--6271. IEEE, 2019.

\bibitem[Hu et~al.(2020)Hu, Anderson, Li, Sun, Carr, Ragan-Kelley, and Durand]{hu2019difftaichi}
Yuanming Hu, Luke Anderson, Tzu-Mao Li, Qi~Sun, Nathan Carr, Jonathan Ragan-Kelley, and Fr{\'e}do Durand.
\newblock Difftaichi: Differentiable programming for physical simulation.
\newblock In \emph{Proceedings of the International Conference on Learning Representations (ICLR)}, 2020.

\bibitem[Jakobi et~al.(1995)Jakobi, Husbands, and Harvey]{jakobi1995noise}
Nick Jakobi, Phil Husbands, and Inman Harvey.
\newblock Noise and the reality gap: The use of simulation in evolutionary robotics.
\newblock In \emph{Proceedings of the European Conference on Artificial Life (ECAL)}, pages 704--720, 1995.

\bibitem[Joachimczak et~al.(2016)Joachimczak, Suzuki, and Arita]{joachimczak2016artificial}
Micha{\l} Joachimczak, Reiji Suzuki, and Takaya Arita.
\newblock Artificial metamorphosis: Evolutionary design of transforming, soft-bodied robots.
\newblock \emph{Artificial Life}, 22\penalty0 (3):\penalty0 271--298, 2016.

\bibitem[Kingma and Ba(2015)]{adamkingma}
Diederik~P. Kingma and Jimmy Ba.
\newblock Adam: {A} method for stochastic optimization.
\newblock In Yoshua Bengio and Yann LeCun, editors, \emph{Proceedings of the International Conference on Learning Representations (ICLR)}, 2015.

\bibitem[Komosi{\'n}ski and Rotaru-Varga(2001)]{komosinski2001comparison}
Maciej Komosi{\'n}ski and Adam Rotaru-Varga.
\newblock Comparison of different genotype encodings for simulated three-dimensional agents.
\newblock \emph{Artificial Life}, 7\penalty0 (4):\penalty0 395--418, 2001.

\bibitem[Koza(1992)]{Koza_1992_Genetic}
John~R. Koza.
\newblock \emph{{Genetic programming: on the programming of computers by means of natural selection}}.
\newblock MIT Press, Cambridge, MA, USA, 1992.

\bibitem[Kr{\v{c}}ah(2010)]{krvcah2010solving}
Peter Kr{\v{c}}ah.
\newblock Solving deceptive tasks in robot body-brain co-evolution by searching for behavioral novelty.
\newblock In \emph{Proceedings of the International Conference on Intelligent Systems Design and Applications}, pages 284--289. IEEE, 2010.

\bibitem[Kriegman et~al.(2018{\natexlab{a}})Kriegman, Cheney, and Bongard]{kriegman2018morphological}
Sam Kriegman, Nick Cheney, and Josh Bongard.
\newblock How morphological development can guide evolution.
\newblock \emph{Scientific Reports}, 8\penalty0 (1):\penalty0 13934, 2018{\natexlab{a}}.

\bibitem[Kriegman et~al.(2018{\natexlab{b}})Kriegman, Cheney, Corucci, and Bongard]{kriegman2018interoceptive}
Sam Kriegman, Nick Cheney, Francesco Corucci, and Josh~C. Bongard.
\newblock Interoceptive robustness through environment-mediated morphological development.
\newblock In \emph{Proceedings of the Genetic and Evolutionary Computation Conference (GECCO)}, pages 109--116. ACM, 2018{\natexlab{b}}.
\newblock URL \url{https://arxiv.org/abs/1804.02257}.

\bibitem[Kriegman et~al.(2020{\natexlab{a}})Kriegman, Blackiston, Levin, and Bongard]{kriegman2020xenobots}
Sam Kriegman, Douglas Blackiston, Michael Levin, and Josh Bongard.
\newblock A scalable pipeline for designing reconfigurable organisms.
\newblock \emph{Proceedings of the National Academy of Sciences}, 117\penalty0 (4):\penalty0 1853--1859, 2020{\natexlab{a}}.

\bibitem[Kriegman et~al.(2020{\natexlab{b}})Kriegman, Nasab, Shah, Steele, Branin, Levin, Bongard, and Kramer-Bottiglio]{kriegman2020scalable}
Sam Kriegman, Amir~Mohammadi Nasab, Dylan Shah, Hannah Steele, Gabrielle Branin, Michael Levin, Josh Bongard, and Rebecca Kramer-Bottiglio.
\newblock Scalable sim-to-real transfer of soft robot designs.
\newblock In \emph{Proceedings of the International Conference on Soft Robotics (RoboSoft)}, pages 359--366, 2020{\natexlab{b}}.

\bibitem[Kriegman et~al.(2021{\natexlab{a}})Kriegman, Blackiston, Levin, and Bongard]{kriegman2021kinematic}
Sam Kriegman, Douglas Blackiston, Michael Levin, and Josh Bongard.
\newblock Kinematic self-replication in reconfigurable organisms.
\newblock \emph{Proceedings of the National Academy of Sciences}, 118\penalty0 (49):\penalty0 e2112672118, 2021{\natexlab{a}}.

\bibitem[Kriegman et~al.(2021{\natexlab{b}})Kriegman, Nasab, Blackiston, Steele, Levin, Kramer-Bottiglio, and Bongard]{kriegman2021fractals}
Sam Kriegman, Amir~Mohammadi Nasab, Douglas Blackiston, Hannah Steele, Michael Levin, Rebecca Kramer-Bottiglio, and Josh Bongard.
\newblock Scale invariant robot behavior with fractals.
\newblock In \emph{Robotics: Science and Systems (RSS)}, 2021{\natexlab{b}}.

\bibitem[Lehman and Stanley(2011)]{lehman2011evolving}
Joel Lehman and Kenneth~O Stanley.
\newblock Evolving a diversity of virtual creatures through novelty search and local competition.
\newblock In \emph{Proceedings of the Genetic and Evolutionary Computation Conference (GECCO)}, pages 211--218, 2011.

\bibitem[Lenski et~al.(2003)Lenski, Ofria, Pennock, and Adami]{lenski2003evolutionary}
Richard~E Lenski, Charles Ofria, Robert~T Pennock, and Christoph Adami.
\newblock The evolutionary origin of complex features.
\newblock \emph{Nature}, 423\penalty0 (6936):\penalty0 139--144, 2003.

\bibitem[Lessin et~al.(2013)Lessin, Fussell, and Miikkulainen]{lessin2013open}
Dan Lessin, Don Fussell, and Risto Miikkulainen.
\newblock Open-ended behavioral complexity for evolved virtual creatures.
\newblock In \emph{Proceedings of the Genetic and Evolutionary Computation Conference (GECCO)}, pages 335--342, 2013.

\bibitem[Li et~al.(2024)Li, Matthews, and Kriegman]{li2024reinforcement}
Muhan Li, David Matthews, and Sam Kriegman.
\newblock Reinforcement learning for freeform robot design.
\newblock In \emph{Proceedings of the International Conference on Robotics and Automation (ICRA)}, 2024.

\bibitem[Lipson and Pollack(2000)]{lipson2000automatic}
Hod Lipson and Jordan~B Pollack.
\newblock Automatic design and manufacture of robotic lifeforms.
\newblock \emph{Nature}, 406\penalty0 (6799):\penalty0 974, 2000.

\bibitem[Ma et~al.(2021)Ma, Du, Zhang, Wu, Spielberg, Katzschmann, and Matusik]{ma2021diffaqua}
Pingchuan Ma, Tao Du, John~Z Zhang, Kui Wu, Andrew Spielberg, Robert~K Katzschmann, and Wojciech Matusik.
\newblock Diffaqua: A differentiable computational design pipeline for soft underwater swimmers with shape interpolation.
\newblock \emph{ACM Transactions on Graphics (TOG)}, 40\penalty0 (4):\penalty0 1--14, 2021.

\bibitem[Matthews et~al.(2023)Matthews, Spielberg, Rus, Kriegman, and Bongard]{matthews2023efficient}
David Matthews, Andrew Spielberg, Daniela Rus, Sam Kriegman, and Josh Bongard.
\newblock Efficient automatic design of robots.
\newblock \emph{Proceedings of the National Academy of Sciences}, 120\penalty0 (41):\penalty0 e2305180120, 2023.

\bibitem[Medvet et~al.(2021)Medvet, Bartoli, Pigozzi, and Rochelli]{medvet2021biodiversity}
Eric Medvet, Alberto Bartoli, Federico Pigozzi, and Marco Rochelli.
\newblock Biodiversity in evolved voxel-based soft robots.
\newblock In \emph{Proceedings of the Genetic and Evolutionary Computation Conference (GECCO)}, pages 129--137, 2021.

\bibitem[Miconi and Channon(2006)]{miconi2006improved}
Thomas Miconi and Alastair Channon.
\newblock An improved system for artificial creatures evolution.
\newblock In \emph{Proceedings of the International Conference on the Simulation and Synthesis of Living Systems (ALife)}, 2006.

\bibitem[Miras et~al.(2020)Miras, Ferrante, and Eiben]{miras2020environmental}
Karine Miras, Eliseo Ferrante, and AE~Eiben.
\newblock Environmental influences on evolvable robots.
\newblock \emph{PloS ONE}, 15\penalty0 (5):\penalty0 e0233848, 2020.

\bibitem[Moreno and Fai{\~n}a(2021)]{moreno2021emerge}
Rodrigo Moreno and Andres Fai{\~n}a.
\newblock {EMERGE} modular robot: a tool for fast deployment of evolved robots.
\newblock \emph{Frontiers in Robotics and AI}, 8:\penalty0 699814, 2021.

\bibitem[Nolfi and Floreano(1998)]{nolfi1998coevolving}
Stefano Nolfi and Dario Floreano.
\newblock Coevolving predator and prey robots: Do ``arms races''' arise in artificial evolution?
\newblock \emph{Artificial Life}, 4\penalty0 (4):\penalty0 311--335, 1998.

\bibitem[Norstein et~al.(2023)Norstein, Veenstra, Ellefsen, Nygaard, and Glette]{norstein2023effects}
Emma~Stensby Norstein, Frank Veenstra, Kai~Olav Ellefsen, T{\o}nnes Nygaard, and Kyrre Glette.
\newblock Effects of compliant and structural parts in evolved modular robots.
\newblock In \emph{Proceedings of the Artificial Life Conference (ALife)}. MIT Press, 2023.

\bibitem[Otto(2009)]{otto2009evolutionary}
Sarah~P Otto.
\newblock The evolutionary enigma of sex.
\newblock \emph{The American Naturalist}, 174\penalty0 (S1):\penalty0 S1--S14, 2009.

\bibitem[Pathak et~al.(2019)Pathak, Lu, Darrell, Isola, and Efros]{pathak2019learning}
Deepak Pathak, Christopher Lu, Trevor Darrell, Phillip Isola, and Alexei~A Efros.
\newblock Learning to control self-assembling morphologies: a study of generalization via modularity.
\newblock In \emph{Advances in Neural Information Processing Systems (NeurIPS)}, 2019.

\bibitem[Pigozzi et~al.(2023)Pigozzi, Medvet, Bartoli, and Rochelli]{pigozzi2023factors}
Federico Pigozzi, Eric Medvet, Alberto Bartoli, and Marco Rochelli.
\newblock Factors impacting diversity and effectiveness of evolved modular robots.
\newblock \emph{ACM Transactions on Evolutionary Learning (TELO)}, 3\penalty0 (1):\penalty0 1--33, 2023.

\bibitem[Rieffel et~al.(2014)Rieffel, Knox, Smith, and Trimmer]{rieffel2014growing}
John Rieffel, Davis Knox, Schuyler Smith, and Barry Trimmer.
\newblock Growing and evolving soft robots.
\newblock \emph{Artificial Life}, 20\penalty0 (1):\penalty0 143--162, 2014.

\bibitem[Schaff et~al.(2022)Schaff, Sedal, and Walter]{schaff2022soft}
Charles Schaff, Audrey Sedal, and Matthew~R Walter.
\newblock Soft robots learn to crawl: Jointly optimizing design and control with sim-to-real transfer.
\newblock In \emph{Robotics: Science and Systems (RSS)}, 2022.

\bibitem[Sims(1994)]{sims1994competition}
Karl Sims.
\newblock Evolving 3{D} morphology and behavior by competition.
\newblock \emph{Artificial Life}, 1\penalty0 (4):\penalty0 353--372, 1994.

\bibitem[Stanley and Miikkulainen(2002)]{stanley2002NEAT}
Kenneth~O Stanley and Risto Miikkulainen.
\newblock Evolving neural networks through augmenting topologies.
\newblock \emph{Evolutionary Computation}, 10\penalty0 (2):\penalty0 99--127, 2002.

\bibitem[van Diepen and Shea(2022)]{van2022co}
Merel van Diepen and Kristina Shea.
\newblock Co-design of the morphology and actuation of soft robots for locomotion.
\newblock \emph{Journal of Mechanical Design}, 144\penalty0 (8):\penalty0 083305, 2022.

\bibitem[Veenstra and Glette(2020)]{veenstra2020different}
Frank Veenstra and Kyrre Glette.
\newblock How different encodings affect performance and diversification when evolving the morphology and control of 2{D} virtual creatures.
\newblock In \emph{Proceedings of the Conference on Artificial Life (ALife)}, pages 592--601, 2020.

\bibitem[Ventrella(1994)]{ventrella1994explorations}
Jeffrey Ventrella.
\newblock Explorations in the emergence of morphology and locomotion behavior in animated characters.
\newblock In \emph{Proceedings of the International Workshop on the Synthesis and Simulation of Living Systems (ALife)}, pages 436--441, 1994.

\bibitem[Wang et~al.(2019)Wang, Zhou, Fidler, and Ba]{wang2019neural}
Tingwu Wang, Yuhao Zhou, Sanja Fidler, and Jimmy Ba.
\newblock Neural graph evolution: Towards efficient automatic robot design.
\newblock In \emph{Proceedings of the International Conference on Learning Representations (ICLR)}, 2019.

\bibitem[Xu et~al.(2021)Xu, Makoviychuk, Narang, Ramos, Matusik, Garg, and Macklin]{xu2021accelerated}
Jie Xu, Viktor Makoviychuk, Yashraj Narang, Fabio Ramos, Wojciech Matusik, Animesh Garg, and Miles Macklin.
\newblock Accelerated policy learning with parallel differentiable simulation.
\newblock In \emph{Proceedings of the International Conference on Learning Representations (ICLR)}, 2021.

\bibitem[Yuan et~al.(2022)Yuan, Song, Luo, Sun, and Kitani]{yuan2022transformact}
Ye~Yuan, Yuda Song, Zhengyi Luo, Wen Sun, and Kris Kitani.
\newblock Transform2{A}ct: Learning a transform-and-control policy for efficient agent design.
\newblock In \emph{Proceedings of the International Conference on Learning Representations (ICLR)}, 2022.

\bibitem[Yuhn et~al.(2023)Yuhn, Sato, Kobayashi, Kawamoto, and Nomura]{yuhn20234d}
Changyoung Yuhn, Yuki Sato, Hiroki Kobayashi, Atsushi Kawamoto, and Tsuyoshi Nomura.
\newblock 4{D} topology optimization: Integrated optimization of the structure and self-actuation of soft bodies for dynamic motions.
\newblock \emph{Computer Methods in Applied Mechanics and Engineering}, 414:\penalty0 116187, 2023.

\bibitem[Zhao et~al.(2020)Zhao, Xu, Konakovi{\'c}-Lukovi{\'c}, Hughes, Spielberg, Rus, and Matusik]{zhao2020robogrammar}
Allan Zhao, Jie Xu, Mina Konakovi{\'c}-Lukovi{\'c}, Josephine Hughes, Andrew Spielberg, Daniela Rus, and Wojciech Matusik.
\newblock Robogrammar: graph grammar for terrain-optimized robot design.
\newblock \emph{ACM Transactions on Graphics (TOG)}, 39\penalty0 (6):\penalty0 1--16, 2020.

\end{thebibliography}

\end{document}